\documentclass[letterpaper,journal]{IEEEtran}

\usepackage{amsmath,amsfonts}
\usepackage{algorithmic}
\usepackage{array}
\usepackage[caption=false,font=normalsize,labelfont=sf,textfont=sf]{subfig}
\usepackage{textcomp}
\usepackage{stfloats}
\usepackage{url}
\usepackage{verbatim}
\usepackage{graphicx}
\usepackage{balance}

\usepackage[utf8]{inputenc}
\usepackage{times}
\usepackage{mathtools}
\usepackage{amssymb}
\usepackage{amsthm}
\usepackage{tabularx}
\usepackage[table]{xcolor}
\usepackage{algorithm}
\usepackage{booktabs}

\usepackage{hyperref} 

\usepackage{xcolor}
\usepackage{comment}
\usepackage[normalem]{ulem}

\usepackage[switch, displaymath, mathlines]{lineno}

\definecolor{xpblue} {rgb}{0.19, 0.55, 0.91}

\definecolor{zqred} {rgb}{0.9, 0.2, 0.2}
\newcommand\ZQ[1]{#1}

\begin{document}

\title{Incremental Micro-Expression Recognition: A Benchmark}

\author{Zhengqin Lai, Xiaopeng Hong, Yabin Wang, and Xiaobai Li%
\thanks{Manuscript received XXX, 2024; revised XXX, 2024. (Corresponding author: Xiaopeng Hong.)}%
\thanks{Z. Lai and X. Hong are with the Harbin Institute of Technology, Harbin, China, and also with Pengcheng Laboratory, Shenzhen, China (e-mail: quenlenu@gmail.com; hongxiaopeng@hit.edu.cn).}%
\thanks{Y. Wang is with Xi'an Jiaotong University, Xi'an, China (e-mail: iamwangyabin@stu.xjtu.edu.cn).}%
\thanks{X. Li is with Zhejiang University, Hangzhou, China (e-mail: xiaobai.li@zju.edu.cn).}%
}

\markboth{IEEE TRANSACTIONS ON AFFECTIVE COMPUTING, VOL. XX, NO. X, JUNE 2024}%
{Lai \MakeLowercase{\textit{et al.}}: Incremental Micro-Expression Recognition: A Benchmark}

\maketitle
\begin{abstract}
Micro-expression recognition (MER) plays a pivotal role in understanding hidden emotions. While traditional methods assume static datasets, real-world scenarios require adapting to continuously evolving data streams. To this end, we introduce the first benchmark specifically designed for Incremental Micro-Expression Recognition (IMER). Our contributions include: Firstly, we formulate a composite class-domain incremental learning setting and construct a sequential benchmark from five representative datasets with carefully curated learning orders to reflect real-world scenarios. Secondly, we establish robust evaluation protocols with a fold-binding strategy to ensure rigorous and feasible cross-session validation, using comprehensive metrics and novel cross-domain visualizations to diagnose performance. Thirdly, we propose Mahalanobis Refinement (MR), a two-stage approach that leverages accumulated second-order statistics for stability and Mahalanobis-constrained refinement for plasticity. Extensive experiments demonstrate that MR significantly outperforms state-of-the-art baselines, effectively balancing the stability-plasticity dilemma. This work lays the foundation for scalable and adaptive micro-expression analysis. All source codes will be released.
\end{abstract}
 
\section{Introduction}

In recent years, we have witnessed a growing interest in micro-expression recognition (MER) due to its critical role in understanding concealed emotions \cite{ekman2003darwin} and its applications in domains such as security and healthcare.
Traditional approaches to MER have predominantly operated assuming that the entire dataset is available at once \cite{karnati2023understanding}. However, real-world scenarios often present a dynamic {environment where data arrives \emph{continuously}}, frequently introducing new categories. This evolving nature of data renders the retraining of models from scratch with each update impractical \cite{masana2022class}, as it entails significant computational overhead and risks catastrophic forgetting \cite{pfulb2019comprehensive}. {Thus, incremental learning, which allows models to adapt progressively to new data while retaining previously learned information \cite{parisi2019continual}, is essential for MER models to address these challenges.}

Despite the necessity of this learning paradigm, it is not trivial to introduce incremental learning to MER directly. Three main challenges need to be addressed.

\begin{figure}[!t]
    \centering
    \includegraphics[width=\linewidth]{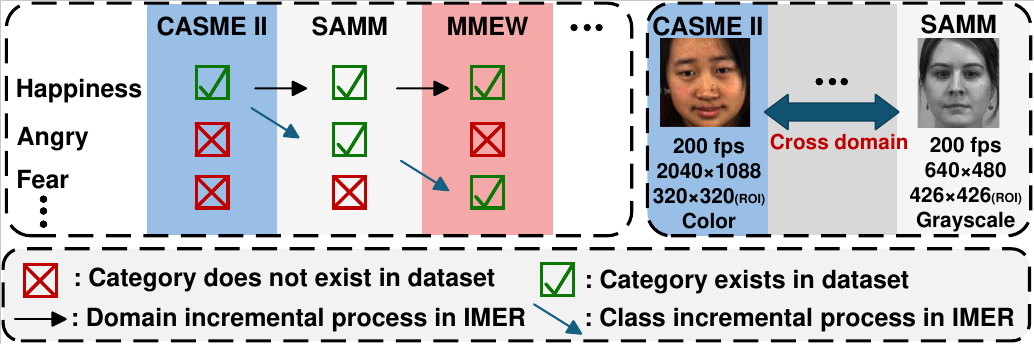}
    \caption{The incremental micro-expression (IMER) task necessitates learning from a mixture of samples belonging to new categories and originating from new domains. \ZQ{For each dataset, the statistics panel lists (from top to bottom) the frame rate, the full-frame resolution, the averaged face-crop size (annotated with ``ROI''), and the color space.}}
    \label{fig:dataset_difference}
\end{figure}

The first challenge lies in \textit{establishing an appropriate learning setup} \textbf{(Challenge 1)}. Typically, incremental learning settings include class-incremental learning \cite{yan2021dynamically,wang2022foster,zhou2022model} and domain-incremental learning \cite{wang2022s,zhu2025reshaping}. In class-incremental learning, models learn new samples from previously unseen classes, whereas in domain-incremental learning, models {adapt to new samples from the same classes but originating from different domains. However, in MER the scenarios are mixed as different micro-expression (ME) datasets contain overlapping but different ME categories. For samples of \emph{seen classes}, the task aligns with \textbf{domain-incremental learning}, where domain differences need to be addressed. As shown in Figure~\ref{fig:dataset_difference}, CASME II and SAMM differ significantly in sensor types (grayscale or RGB), image resolutions, and subject demographics. Furthermore, datasets often use different protocols for inducing MEs. For samples of \emph{unseen classes}, the task transitions to \textbf{class-incremental learning}. {For example, when sequentially learning from the CASME II, SAMM, and MMEW datasets, \emph{Angry} in SAMM and \emph{Fear} in MMEW emerge as new categories.}  Consequently, each learning phase involves a combination of samples from new categories and new domains.
 This unique challenge means that existing incremental learning paradigms cannot address the practical scenarios encountered in IMER.}

\textit{Organizing data} poses another significant obstacle \textbf{(Challenge 2)}. Traditional incremental learning studies\cite{li2023continual} typically segment a single dataset into sequential sessions, e.g., the CIFAR-100 benchmark is divided into ten sessions of ten distinct classes (commonly referred to as the `10-10' setting). However, such \emph{single-dataset-split} protocols are infeasible for MER. ME datasets typically consist of only a few categories and several hundred samples, which are insufficient for further splitting.

{Finally, a critical issue lies in \textit{evaluating the cross-subject generality} \textbf{(Challenge 3)}. Typical incremental learning studies usually do not consider cross-subject evaluation, which is an important concern in (micro-) expression recognition. One straightforward approach is to perform the cross-subject evaluation in each learning session and then perform a combined evaluation across all sessions. However, this would lead to a \emph{combinatorial explosion} of evaluation trials. For example, if there are 10, 12, and 15 subjects in three consecutive sessions, a direct combination would result in $10 \times 12\times 15 =$ 1800 trials to obtain a single performance metric value, which is impractical. }

\ZQ{To address these challenges, our work makes three primary contributions.}

\ZQ{First, to tackle the difficulties in learning setup and data organization (\textbf{Challenge 1} and \textbf{Challenge 2}), we establish a foundational benchmark for IMER. We formulate a composite class-domain incremental learning setting to model the unique mixed nature of MER tasks. Correspondingly, we construct a sequential benchmark from five representative datasets, organized chronologically to simulate real-world data streams.}

\ZQ{Second, to resolve the challenge of evaluating cross-subject generality (\textbf{Challenge 3}), we establish robust evaluation protocols featuring a novel fold-binding strategy. This strategy effectively regulates the cross-session validation process, avoiding the combinatorial explosion of evaluation trials while ensuring rigorous performance assessment. Additionally, we provide comprehensive metrics to diagnose performance, specifically considering the impact of class imbalance.}

\ZQ{Third, to address the stability and plasticity dilemma revealed by our benchmark, we propose Mahalanobis Refinement (MR), a novel two-stage approach. To ensure stability, MR constructs a general evolution model from accumulated second-order statistics, creating a robust Mahalanobis metric space that implicitly de-correlates features. To enhance plasticity, it performs local refinement using a margin-based loss, but critically, this process is constrained by the Mahalanobis distance to preserve the learned structure.}

\ZQ{The remainder of this paper is organized as follows. Section II reviews related work in MER and incremental learning. Section III details the construction of our IMER benchmark, including problem definition, data organization, and evaluation protocols. Section IV presents our proposed Mahalanobis Refinement (MR) method. Section V and VI report the experimental results and discussions. Finally, Section VII concludes the paper.}
 
\section{Related Work}

\subsection{Micro-Expression Recognition (MER)}
\label{sec:mer_related}
The evolution of MER has witnessed a shift from traditional approaches to neural network-based methods.

Early studies relied on traditional methods like Local Binary Patterns~\cite{li2013spontaneous} and LBP-TOP~\cite{pfister2011recognising} for texture feature extraction. The deep learning era emerged around 2016 with CNN-based feature transfer~\cite{patel2016selective}, followed by CNN-LSTM hybrid models~\cite{kim2016micro} that combined spatiotemporal analysis. Further developments included CNN-based optical flow processing~\cite{gan2019off} and Recurrent CNNs~\cite{gan2019off}, significantly outperforming traditional approaches.

Recent transformer-based innovations have achieved remarkable progress. Notable examples include SLSTT~\cite{zhang2022short}, which combines transformers with LSTMs, the Muscle Motion-guided Network~\cite{li2022mmnet} featuring continuous attention modules, and Micron-BERT~\cite{nguyen2023micron} that introduces self-supervised pre-training for enhanced emotion detection. \ZQ{More recently, advanced techniques involving graph convolutions and attention mechanisms have been explored for joint estimation of MEs and AUs~\cite{shao2025mol,shao2025facial,shao2026constrained}.}

While these methods have demonstrated impressive performance, they predominantly operate under a \emph{batch learning} paradigm. \ZQ{Although some works utilize cross-dataset or composite-dataset protocols (e.g., the MEGC2019 protocol~\cite{xia2020revealing}), they typically focus on static domain generalization rather than incremental adaptation.} Such approaches may not fully address real-world scenarios where continuous learning from new data streams is essential.

\subsection{Datasets for MER}

Emerging datasets have significantly advanced the study of MER. Well-known datasets such as SMIC~\cite{smic}, SAMM~\cite{samm}, the CASME series (CASME~\cite{yan2013casme}, CASME II~\cite{yan2014casme}, CAS(ME)$^3$~\cite{li2022cas3}), MMEW~\cite{mmew}, and 4DME~\cite{li20224dme} offer detailed annotations including onset, apex, and offset frames, along with Action Unit (AU)~\cite{ekman1978facial} labels. One of the latest ME datasets, CAS(ME)$^3$, offers extensive multimodal data of a larger sample size, increasing the ecological validity for researchers. \ZQ{Furthermore, DFME~\cite{zhao2023dfme}, the largest spontaneous ME dataset to date, provides over 7,000 samples with high frame rates, significantly alleviating the data scarcity issue.} However, most previous ME datasets contain fewer than 300 samples. Given the limited size of individual datasets, splitting a single dataset for incremental learning experiments is impractical. Instead, a cross-dataset protocol following the temporal order is more appropriate for incremental MER.

\subsection{Incremental Learning}

Incremental learning is a paradigm in machine learning that enables models to learn continuously from new data while retaining previously acquired knowledge. This approach addresses the challenges posed by real-world applications where data becomes available sequentially, and models need to adapt without retraining from scratch, which is computationally expensive and risks catastrophic forgetting~\cite{mccloskey1989catastrophic,mcclelland1995there}.

\textbf{Class-Incremental Learning (CIL)} focuses on the continual addition of new classes, where the primary challenge is to recognize both new and previously learned classes simultaneously. Early methods like Learning without Forgetting~\cite{li2017learning} utilized knowledge distillation to preserve prior knowledge. iCaRL~\cite{rebuffi2017icarl} introduced a strategy combining representation learning with a nearest-mean-of-exemplars classifier and maintained a memory buffer of exemplars from previous classes to mitigate forgetting. Other notable methods include DER~\cite{yan2021dynamically}, which dynamically expands the model's capacity by adding new modules for new sessions while preserving previous parameters, and FOSTER~\cite{wang2022foster}, which proposes a two-stage learning paradigm that first dynamically expands new modules and then compresses the model to maintain a single backbone through an effective distillation strategy. Recent approaches have also leveraged prompt-based learning with pre-trained models. For instance, L2P~\cite{wang2022learning} and DualPrompt~\cite{wang2022dualprompt} learn and manage a pool of prompts to guide the model on new sessions. RanPAC~\cite{mcdonnell2023ranpac} enhances feature separation through random projections between pre-trained features and the output layer.

\textbf{Domain-Incremental Learning (DIL)} addresses the adaptation of models to new data domains or distributions over time. The key challenge is handling concept drift~\cite{kirkpatrick2017overcoming,wang2021concept, wang2022s}, where the underlying data distribution changes, without forgetting previously learned domains. Methods in this area often focus on learning domain-invariant features or adapting models to new domains. For example, S-Prompt~\cite{wang2022s} learns domain-specific prompts on a pre-trained model, avoiding the need to rehearse previous data.

While these settings have been extensively studied, our MER scenario presents unique challenges by inherently combining both class and domain incremental aspects, as MEs can vary across different domains while new ME categories may emerge. Conventional incremental learning settings and their corresponding methods, for either class or domain incremental learning separately, are not directly applicable and require substantial adaptation for this hybrid scenario.

\section{Benchmark Construction}

\subsection{Problem Definition.} 
\label{sec:problem_definition}

{Consider a sequence of datasets $D^{(1)}, D^{(2)}, \dots,$ $D^{(n)}$, where each dataset $D^{(t)} = \{(x_j^t, y_j^t, id_j^t)\}_{j=1}^{N_t}$ consists of micro-expression samples, their corresponding emotion labels, and subject identifiers. Here, $N_t$ denotes the number of samples in the $t$-th dataset. Let $l^{(t)}$ be the set of emotion classes in $D^{(t)}$, and define the cumulative emotion label set up to $t$ as $L_{t} = \bigcup_{k=1}^t l^{(k)}$. Our goal is to learn a mapping function from the sample space $\mathbb{X}$ to their emotion label space $\mathbb{Y}$, using subject identifiers $id$ for data splitting in the evaluation protocol.
The model $\Theta$ is incrementally trained on the data sequence, where each $D^{(t)}$ is available only during its respective training session. At each session, $\Theta$ is updated sequentially while expanding its recognition scope from $L_{t-1}$ to $L_{t}$. After training on $D^{(t)}$,  the updated model $\Theta_t$ is expected to recognize \emph{all} emotion classes encountered  in $L_{t}$.}

Unlike CIL, where $l^{(t)}$ are disjoint, and DIL, where $l^{(t)}$ are identical, IMER presents a \emph{composite class-domain incremental learning} scenario, where label sets across sessions can overlap but differ, namely $l^{(t)} \cap l^{(s)} \neq \emptyset$ and $l^{(t)} \neq l^{(s)}$, as datasets $D^{(t)}$ introduce both novel classes and new domain samples for existing ones. \ZQ{For instance, when transitioning from CASME II to SAMM, the model must adapt to sensor differences for shared classes like \emph{Happiness} (DIL) while simultaneously learning the new class \emph{Anger} (CIL).}

\subsection{Data, Evaluation Protocol, and Metric}
\label{sec:data_protocol}
\begin{figure}[!t]
\centering
\includegraphics[width=\linewidth]{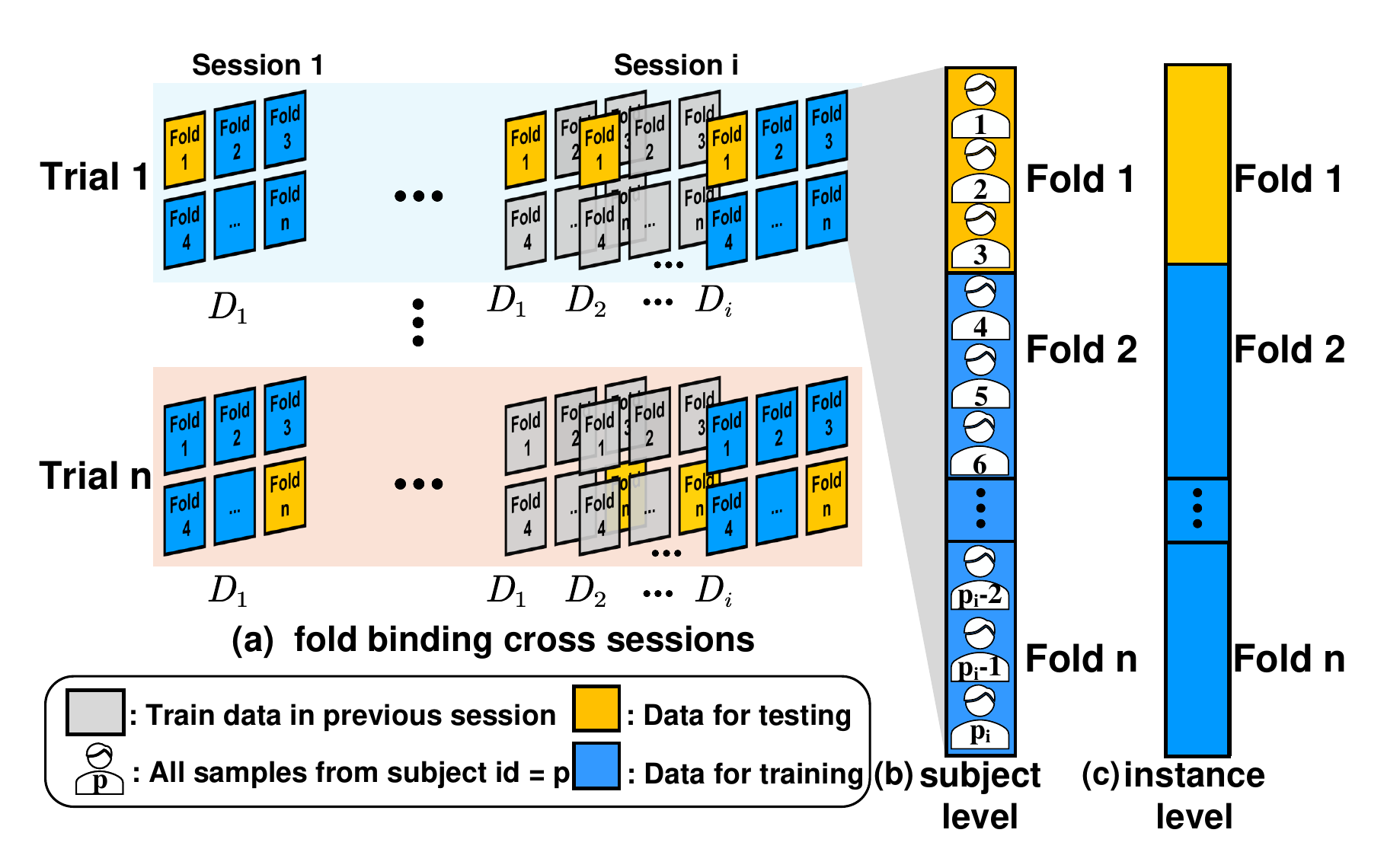}
\caption{Illustration of the evaluation protocols. (a) Fold binding for cross-session evaluation, where the same fold index is {aligned} across sessions, and each session uses all previously bound folds and the current trial-bound fold as test sets (shown in yellow). (b) and (c) are the within-session data partition protocols in the pipeline of (a). In (b), the subject level partition strategy, data is divided based on subject identifiers; In (c), the instance level partition strategy, individual samples are randomly distributed across folds. 
}
\label{fig:Protocol}
\end{figure}
Establishing appropriate data organization and evaluation protocols is crucial yet challenging for  {initiating the study of IMER. The limited size of individual MER datasets renders conventional single-dataset splitting inappropriate. Therefore, we organize the data sequence for incremental learning experiments in a cross-dataset manner, according to the chronological order of dataset publication dates as listed in Table~\ref{table:dataset_sequence}. Later datasets typically encompass a broader range of categories and pose more challenging sessions. Conducting IMER in this sequential order better simulates the iterative versioning process of practical systems while also introducing significant challenges due to the pronounced domain gaps between datasets~\cite{roth2024practitioner}.}

\ZQ{We selected five representative and publicly available datasets (CASME II, SAMM, MMEW, CAS(ME)$^3$, and DFME) to construct the incremental sessions. These datasets utilize standard 2D video formats and provide granular emotion annotations, ensuring a consistent and robust evaluation pipeline. Datasets like SMIC and 4DME were excluded due to incompatible annotation granularity (e.g., coarse 3-class or grouped negative emotions), which would undermine the fine-grained evaluation.
Crucially, our benchmark is designed to be \emph{extensible}. We adopt a unified dataset interface together with a class-mapping file for projecting dataset-specific labels into the unified label space. With this interface, future datasets can be integrated as new sessions by converting their annotations into the specification, without modifying the benchmark pipeline.}

\begin{table}[!t]
\centering
\small
\caption{Dataset Sequence and Related Information}
\label{table:dataset_sequence}
\begin{tabularx}{\linewidth}{@{} l l c >{\raggedright\arraybackslash}X @{}}
\toprule
\textbf{Session} & \textbf{Dataset} & \textbf{Year} & \multicolumn{1}{c}{\textbf{Emotion Classes}} \\
\midrule
Session 1 & CASME II & 2014 & disgust, happiness, others, repression, surprise \\ 
Session 2 & SAMM$^*$ & 2018 & anger, contempt, happiness, others, surprise \\ 
Session 3 & MMEW & 2022 & disgust, fear, happiness, others, sad, surprise \\ 
Session 4 & CAS(ME)$^3$ & 2023 & anger, disgust, fear, happiness, others, sad, surprise \\
\ZQ{Session 5} & \ZQ{DFME} & \ZQ{2023} & \ZQ{anger, contempt, disgust, fear, happiness, sad, surprise} \\
\bottomrule
\end{tabularx}
\begin{flushleft}
\footnotesize *Following standard practice\cite{zhang2022short, xia2020revealing, Lei_2021_CVPR}, we exclude categories with fewer than 10 samples.

\end{flushleft}
\end{table}

\subsubsection{Evaluation Protocol}
\label{sec:evaluation_protocol}

{Traditional MER typically adopts n-fold cross-validation protocols, such as LOSO (Leave-One-Subject-Out). However, in incremental learning, cross-validation must be performed in a cross-session manner, referred to as Cross-Session Validation (CSV). As discussed in the introduction, directly applying k-fold CV across multiple sessions can lead to a combinatorial explosion in the number of validation trials. To address this, we propose a strategy called fold binding to constrain the complexity of CSV. Specifically, each dataset in an incremental session is divided into k segments, and segments with the same fold index across different sessions are bound together. In the $\tau$-th validation trial for $\tau \in \{1,\cdots,k\}$, the $\tau$-th segment from each session is used for testing, while the remaining segments are used for training. This fold binding mechanism ensures regulated cross-session evaluation, reducing the overall number of validation trials to $O(k)$, as illustrated in Figure~\ref{fig:Protocol} (a). For \emph{within-session} experiments, we adopt two cross-validation protocols. In \textbf{Subject Level Cross Validation (SLCV)}, data is partitioned such that all samples from a given subject belong to only one fold. In contrast, \textbf{Instance Level Cross Validation (ILCV)} randomly splits data segments regardless of subject identity.} These protocols offer complementary perspectives on model evaluation. \ZQ{We designate SLCV as the primary protocol because it rigorously assesses the model's ability to generalize across different subjects, a critical requirement for real-world applications. ILCV serves as a supplementary protocol, evaluating the model's overall learning capability on randomly sampled instances without strict subject constraints.} In our experiments, $k=5$ for both protocols.

\subsubsection{Evaluation Metrics}
\label{sec:evaluation_metrics}

In incremental learning, model evaluation requires assessing the performance in both the current session and all previous sessions.

The accuracy \( A_i \) of a single session $i$  is defined as $A_i = \frac{C_i}{T_i}$, where $C_i$ denotes the number of correctly classified samples and $T_i$ represents the total number of test samples for \emph{all} encountered classes in $L_i$. \ZQ{Additionally, considering the class imbalance inherent in MER datasets, we also employ the Unweighted Average Recall (UAR) and F1 score as complementary metrics. The UAR for session $i$, denoted as $U_i$, is calculated as the macro-average of recall across all classes: $U_i = \frac{1}{|L_i|} \sum_{c \in L_i} \frac{C_{i,c}}{N_{i,c}}$, where $C_{i,c}$ is the number of correctly classified samples for class $c$, and $N_{i,c}$ is the total number of samples for class $c$ in the test set. The F1 score for session $i$, denoted as $F1_i$, is calculated as the macro-average of F1 scores across all classes.}

On this basis, cross-session incremental learning performance is evaluated \ZQ{for each of the $k$ folds. The final reported results are averaged over these $k$ folds to ensure reliable evaluation}. We use both the final performance and average performance \cite{zhou2023revisiting} across all sessions. {The \emph{final performance} (denoted as \( \tilde{A} \) for accuracy) reflects the model's overall performance after the last session (e.g., $\tilde{A} = A_n$). The \emph{average performance} (denoted as \( \bar{A} \) for accuracy, \( \bar{U} \) for UAR, and \( \bar{F}_1 \) for F1 score) reflects the progressive performance by averaging the metrics at each session (e.g., $\bar{A} = \frac{1}{n} \sum_{i=1}^{n} A_i$).}

\subsection{Baseline Methods}

To provide the baseline methods for the IMER study, we {introduce a pipeline to upgrade typical MER methods to their incremental learning counterparts, as shown in Fig.~\ref{fig:adapt}.}

\subsubsection{Selection and Adaptation of Base Models}
\label{sec:base_models} 

MER is essentially a video classification task. However, previous research has demonstrated that utilizing only key frames can achieve competitive performance~\cite{liong2018less,li2022mmnet,nguyen2023micron,zhai2023feature}. Additionally, optical flow-based methods~\cite{zhang2022short} have also proven effective in capturing the subtle facial movements characteristic of MEs. Following this established direction in the MER pipeline, we build our benchmark upon three widely-used backbones: ResNet~\cite{he2016identity}, Vision Transformer (ViT)~\cite{dosovitskiy2020image}, and Swin Transformer (SwinT)~\cite{liu2021swin}. For model input, we use the optical flow map between the apex and onset frames, with the face region cropped based on landmarks. These selected base models serve as the foundation for incorporating incremental learning capabilities. We validated {our implementation of} these models to ensure their single dataset performance aligns with state-of-the-art approaches, as shown in the experimental section. 

\subsubsection{Enabling Incremental Learning and Update}
{We embed the base models into modern incremental learning (IL) frameworks. Three representative IL strategies are used, including pre-trained models with prompting, such as L2P~\cite{wang2022learning} and DualPrompt~\cite{wang2022dualprompt}; dynamic network methods like DER~\cite{yan2021dynamically} and FOSTER~\cite{wang2022foster}; and class-prototype-based approaches exemplified by  RanPAC~\cite{mcdonnell2023ranpac}.}

\subsubsection{Unifying and Consolidating Output}
\label{sec:remappable_head_detail} 

\begin{figure}[!t]
    \centering
    \includegraphics[width=1\columnwidth]{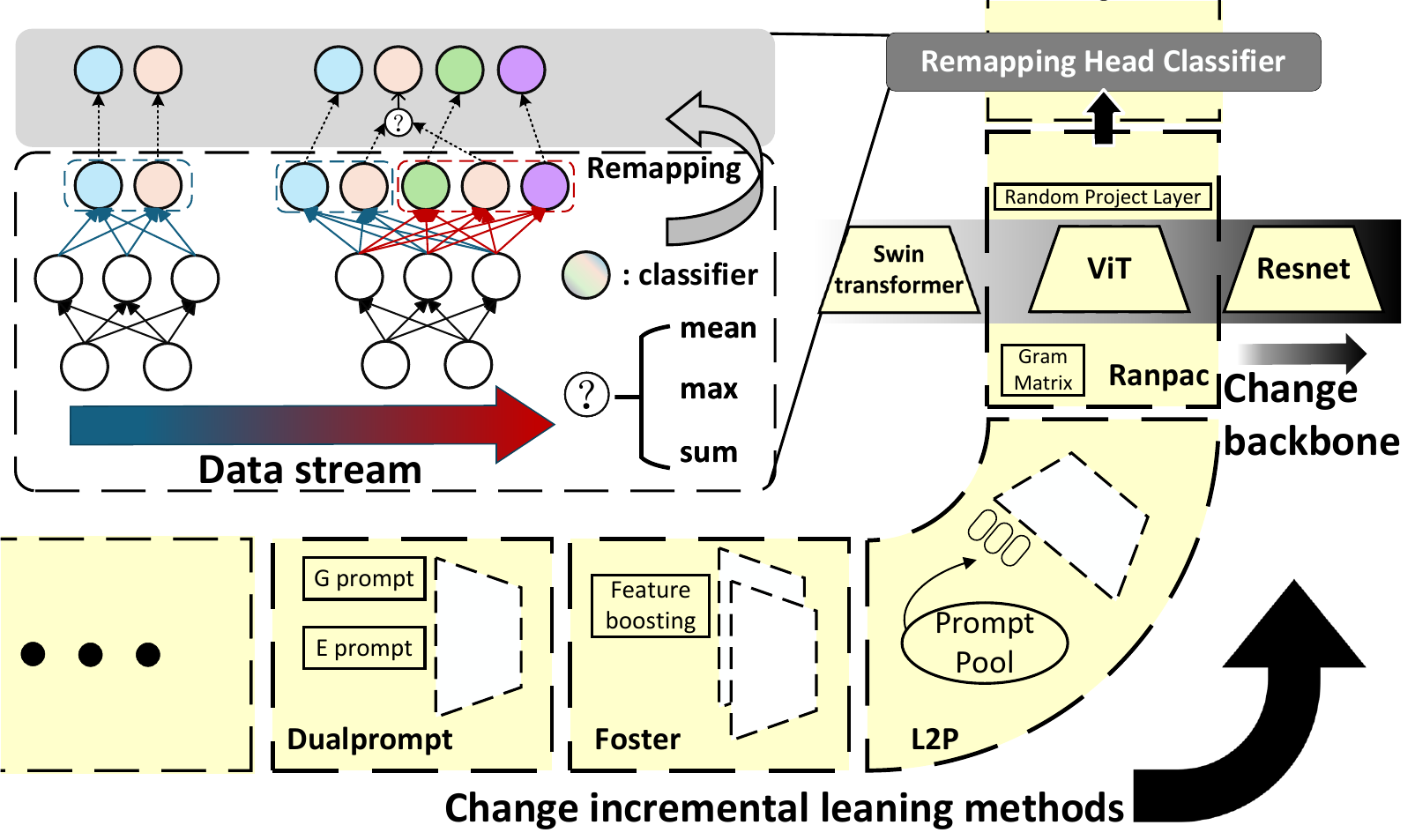}
\caption{{Remappable Classification Head (a) and pipeline (b) for making normal MER approaches incremental. Various backbone structures and incremental learning strategies can be integrated. The remappable classification head mechanism consolidates the outputs by maintaining unified, compact classification heads.}}

    \label{fig:adapt}
\end{figure}

As introduced previously, an ME category may correspond to multiple heads for different sessions. To address this issue, we propose a Remappable Classification Head (RCH) mechanism to maintain compact classification heads (Figure~\ref{fig:adapt} (a)).

We first dynamically {concatenate} session-specific classification heads during training, permitting a single ground-truth class to be represented by multiple distinct heads corresponding to the different sessions in which it appeared. Consequently, during inference, consolidating the outputs (logits $z^{(t)}_c(x)$ for input $x$, class $c$, session $t$) from these diverse session-specific heads becomes necessary to obtain a final prediction for each ground-truth class. We explored three remapping strategies for aggregating the logits for a ground-truth class $c$: the \emph{Max} operation, which selects the maximum logit across relevant sessions ($z_{c, \text{max}}(x) = \max_{t \in T_c} z^{(t)}_c(x)$ where $T_c$ is the set of sessions containing class $c$); the \emph{Mean} operation, calculating the average logit ($z_{c, \text{mean}}(x) = \frac{1}{|T_c|} \sum_{t \in T_c} z^{(t)}_c(x)$); and the {\emph{Sum} operation}, summing the logits ($z_{c, \text{sum}}(x) = \sum_{t \in T_c} z^{(t)}_c(x)$).

The experimental results (averaged across all baseline methods, Table \ref{tab:remapping_results}) confirm the superior performance of the \emph{Max} strategy ($\tilde{A}$ and $\bar{A}$). Its effectiveness lies in selecting the strongest prediction signal for a class across sessions, ensuring robustness to domain shifts, unlike \emph{Mean} or \emph{Sum} which can dilute strong signals. Critically, analysis shows that \emph{Max} remapping is mathematically equivalent to selecting the globally highest logit across all session-specific heads. This is because the globally maximum logit must, by definition, be greater than or equal to the logit of any other class in any session. Consequently, the class that produces this global maximum will also have the highest value after the \emph{Max} operation is applied to it, ensuring it remains the winning class. Thus, both strategies are guaranteed to yield the same final prediction. This equivalence demonstrates that the Remappable Classification Head (RCH) mechanism with \emph{Max} remapping achieves optimal classification while enabling a consolidated, compact head structure. Furthermore, since RCH uses a parameter-free aggregation strategy and fixed after training, its application during inference adds negligible overhead.

\begin{table}[!t]
\centering
\small
\setlength{\tabcolsep}{4pt}
\caption{Average ($\bar{A}$) and final ($\tilde{A}$) accuracy (\%)  for ILCV and SLCV under different aggregation strategies.}
\begin{tabular}{lcccc}
\toprule
 {$\bar{A}$/$\tilde{A}$}& Max & Mean & Sum & None \\
\midrule
ILCV & \textbf{36.40/31.22} & 30.02/22.87 & 34.74/27.60 & 28.18/18.10 \\
SLCV & \textbf{33.09/30.16} & 27.05/21.96 & 31.82/26.90 & 24.35/16.94 \\
\bottomrule
\end{tabular}
\label{tab:remapping_results}
\end{table}

\section{Our Method}

Incremental Micro-Expression Recognition (IMER) faces the critical challenge of balancing stability and plasticity\cite{ditzler2015learning, mermillod2013stability, grossberg2013adaptive}. 
{Previous incremental learning approaches usually rely on first-order statistics (feature means) to maintain and update the classifiers, potentially limiting their ability to capture complex data distributions inherent in MEs. Encouraged by the merits of the second-order statistics in capturing feature correlation~\cite{hong2009sigma, pourkeshavarzi2021looking} and regulating features~\cite{pourkeshavarzi2021looking}, we build our incremental MER approach on top of second-order statistics. Our method, termed Mahalanobis Refinement (MR), is composed of two main components.}

The first, the General Evolution Model (GEM), establishes stability by accumulating second-order statistics across sessions to create a robust general foundation and a Mahalanobis metric space for knowledge preservation. The second, Local Adaptation, enhances plasticity and discriminability by refining the model on current session data using techniques like margin-based loss, while crucially employing a Mahalanobis distance constraint derived from the first stage to maintain stability during adaptation.

\subsection{General Evolution Model Construction}
\label{sec:general_evolution}

Our approach builds upon linear classification models where a feature vector $x \in \mathbb{R}^d$ is classified based on $x^TW$. To lay the groundwork, we first analyze this model in a standard, non-incremental setting. Here, $W \in \mathbb{R}^{d \times K}$ contains the classification weights for all $K$ classes. While various optimization objectives exist for learning $W$, we adopt the regularized least-squares framework. This choice is motivated by its closed-form solution, its natural decomposition into accumulable statistics, and its inherent connection to Mahalanobis distances, properties that prove particularly valuable in an incremental learning context.

Given a set of training features $X \in \mathbb{R}^{N \times d}$ and their corresponding one-hot labels $Y \in \mathbb{R}^{N \times K}$, the objective is to find the weight matrix $W$ that minimizes:
\begin{equation}
L_{LS} = \frac{1}{2} || XW - Y ||_F^2 + \frac{\lambda}{2} ||W||_F^2,
\end{equation}
where $\lambda$ is the L2 regularization parameter. This optimization problem admits a well-known closed-form solution:
\begin{equation}
W = (X^TX + \lambda I)^{-1}X^TY = M^{-1}H,
\label{eq:closed_form_w} 
\end{equation}
where $M = X^TX + \lambda I \in \mathbb{R}^{d \times d}$ is the regularized second-order moment matrix, and $H = X^TY \in \mathbb{R}^{d \times K}$ is a matrix linking first-order statistics with their corresponding labels. \ZQ{Intuitively, $M$ captures feature correlations as a regularized second-order moment matrix (an uncentered covariance estimate up to scale), while $H$ aggregates the feature sums for each class. Consequently, $M^{-1}$ is a symmetric positive definite precision matrix that induces a Mahalanobis metric. Equivalently, for a whitening transform $G$ satisfying $G^\top G=M^{-1}$ (e.g., $G=M^{-1/2}$), the class score can be written as $x^\top w_k=x^\top M^{-1}h_k=(Gx)^\top(Gh_k)$, i.e., a standard inner product similarity after whitening~\cite{de2000mahalanobis,bishop2006pattern,kulis2013metric}.} Inference is performed by computing the scores $x^TW$, with the predicted class being the one corresponding to the highest score.

The key to adapting this formulation to an incremental learning scenario lies in the decomposable nature of the statistics $M$ and $H$, which allows the model to be updated sequentially. Thus, to construct a general model capable of reflecting the evolving trends across all seen data, we accumulate both the second-order moment matrix $M$ and the first-order association matrix $H$. The second-order moment matrix is incrementally accumulated as:
\begin{equation}
M_c^{t} = M_c^{t-1} + M^t,
\label{eq:m_c} 
\end{equation}
where $M^t = (X^t)^TX^t+\lambda I$. The first-order association matrix $H$ is accumulated as:
\begin{equation}
H_c^{t} = [H_c^{t-1}, H^t],
\label{eq:h_c} 
\end{equation}
where $H^t = (X^t)^TY^t$. It is worth noting how the shapes of these matrices evolve. The accumulated second-order moment matrix, $M_c^t$, consistently maintains its dimensions of $d \times d$, where $d$ is the feature dimensionality, as it only accumulates statistical information within the same feature space. In contrast, the first-order association matrix, $H_c^t$, expands its width with each new session to accommodate new classes. A more detailed analysis of parameter growth will be discussed later. We then compute a single, unified set of prototypes optimized considering the general evolution of statistics:
\begin{equation}
W^{t} = (M_c^{t})^{-1} H_c^{t}.
\label{eq:final_wt} 
\end{equation}

This formulation (Eq. \ref{eq:final_wt}) is the core of the general evolution model. It allows for the joint optimization and refinement of all class prototypes at each learning session by leveraging the accumulated statistics. This update mechanism is particularly important for MER, where classes are highly correlated. While fixed class prototypes can be effective when inter-class correlation is low (e.g., cars vs. tigers), we find that allowing previous prototypes to be subtly adjusted when learning new ones enhances feature separation and improves model stability. Our accumulation of both second-order statistics ($M_c^t$) and first-order associations ($H_c^t$) achieves this, enabling the model to learn robust feature relationships while permitting nuanced adjustments to class representations.
\begin{figure}[!t]
\centering
\includegraphics[width=1\columnwidth]{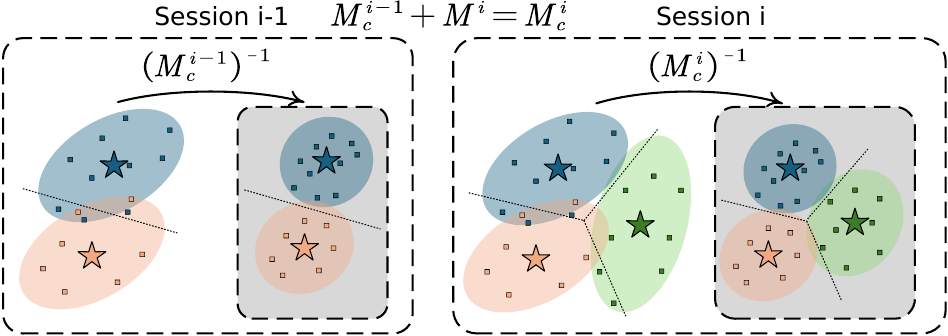}
\caption{By accumulating the $M_c$ matrix over sessions, the general evolution model implicitly computes prototype-sample similarity under a Mahalanobis metric, yielding enhanced decision boundaries within each session.}
\label{fig:overall} 
\end{figure}

Crucially, this general evolution model utilizes the accumulated matrix $M_c^t$. Its inverse, $(M_c^{t})^{-1}$, used in classifier decisions ($x^T W^{t} = x^T (M_c^{t})^{-1} H_c^{t}$), induces a Mahalanobis metric for similarity measurement that accounts for the feature correlations accumulated across all sessions, as illustrated in Figure~\ref{fig:overall}. Operating within this metric space ensures stable knowledge inheritance and continuous modeling of the general evolution.

\subsection{Local Adaptation }
\label{sec:local_adaptation}

While the general evolution model provides excellent stability and knowledge continuity through accumulated statistics, its foundation on the least-squares objective (L2 loss), which minimizes regression error, does not directly maximize classification margins. This may impose a theoretical limitation on its final discriminative ability, especially when distinguishing highly similar or subtle patterns like MEs.

To overcome this limitation and further enhance the model's local adaptivity and discriminative power, we introduce a local adaptation and refinement module. This module takes the weights $W^t$ provided by the general evolution model as an informed initialization (referred to as $W_{init}$) and performs fine-tuning using the data from the current session $(X^t, Y^t)$.

We employ gradient-based deep learning optimization for this refinement step. Regarding the loss function, while standard Cross-Entropy (CE) loss is an option, a margin-based loss function is more effective for maximizing inter-class separability, particularly for fine-grained recognition tasks like MEs. Therefore, we adopt the ArcFace Loss \cite{deng2019arcface}. ArcFace enhances class separation by introducing an additive angular margin $m$ to the target angle $\theta_{y_i}$ in the angular space:
\begin{equation}
 L_{a} = -\frac{1}{N_t} \sum_{i=1}^{N_t} \log \frac{e^{s(\cos(\theta_{y_i}+m))}}{e^{s(\cos(\theta_{y_i}+m))} + \sum_{j=1, j \neq y_i}^{K_t} e^{s\cos\theta_j}},
 \label{eq:arcface_loss} 
\end{equation}
where $s$ is a scaling factor, $N_t$ is the number of samples in the current session, and $K_t$ is the total number of classes encountered up to session $t$. This encourages greater inter-class distance and intra-class compactness in the angular space, aligning well with our cosine similarity-based classifier initialized by $W_{init} = W^t$.

However, naive fine-tuning with ArcFace alone can lead to instability and disrupt the learned prototype structure as shown in Section~\ref{sec:AMR}. To ensure stability during this plasticity enhancement, we introduce a novel regularization term based on the Mahalanobis distance.

This term directly enforces stability by penalizing deviations of the refined weights $W$ from the initialized weights $W_{init}$, measured within the Mahalanobis space defined by the accumulated statistics $M_c^t$ from the general evolution model. Starting from constraining the Mahalanobis distance between initial ($H_{init}=M_c^t W_{init}$) and updated ($H=M_c^t W$) prototypes, specifically $\text{tr}((H - H_{init})^T (M_c^t)^{-1} (H - H_{init}))$, we arrive at the regularization term expressed in terms of weights. \ZQ{From Eq.~\ref{eq:final_wt}, we have:}

\begin{align}
L_m &= \text{tr}((M_c^t W - M_c^t W_{init})^T (M_c^t)^{-1} (M_c^t W - M_c^t W_{init})) \nonumber  \\
    &= \text{tr}((W - W_{init})^T M_c^t (W - W_{init})). \label{eq:mahalanobis_regularization}
\end{align}

Conceptually, the Mahalanobis constraint (Eq. \ref{eq:mahalanobis_regularization}) ensures that local refinement respects the general structure learned across previous sessions, anchoring the adaptation process by leveraging the metric derived from the general evolution model.

Our final loss function for local adaptation synergistically combines the discriminative ArcFace loss with the stabilizing Mahalanobis regularization:
\begin{equation}
 L = L_{a} + \alpha \cdot L_m = L_{a} + \alpha \cdot \text{tr}((W - W_{init})^T M_c^t (W - W_{init})),
 \label{eq:combined_loss}
\end{equation}
where $\alpha$ is a hyperparameter balancing the two terms.

This Mahalanobis-constrained refinement allows the model to effectively learn new information and enhance discriminability (plasticity) using a powerful margin-based loss, while the constraint, rooted in the general evolution model's metric, prevents catastrophic forgetting (stability). 

In essence, our two-stage MR approach, as summarized in Algorithm \ref{alg:our_method}, achieves the desired balance: GEM provides stability through accumulated statistics and a Mahalanobis metric, while the Local Adaptation stage enhances plasticity via Mahalanobis-constrained refinement using the combined loss (Eq. \ref{eq:combined_loss}). The backbone $\phi$ is first adapted on the initial data session $D^{(1)}$ to bridge the domain gap between the pre-training dataset and the target ME domain.

\begin{algorithm}[!t]
\caption{Mahalanobis Refinement (MR)}
\label{alg:our_method}
\textbf{Input}: Sequential datasets $\{D^{(t)}\}_{t=1}^n$; An adapted backbone $\phi$; Hyperparameters $\lambda, \alpha$. \\
\textbf{Output}: Final model $(\phi, W_n)$.
\begin{algorithmic}[1]
\STATE Initialize $M_c \leftarrow \mathbf{0}$, $H_c \leftarrow \mathbf{0}$.
\FOR{session $t = 1$ to $n$}
    \STATE Extract features $F^t \leftarrow \phi(X^t)$.
    \STATE Update statistics $M_c$ and $H_c$ by Eq. (\ref{eq:m_c}) and (\ref{eq:h_c}).
    \STATE Compute initial prototypes $W_{t}$ by Eq. (\ref{eq:final_wt}). //General evolution model.
    \STATE Refine model by minimizing loss $L$ in Eq. (\ref{eq:combined_loss}). //Local adaptation.
\ENDFOR
\STATE \textbf{return} $(\phi, W_n)$.
\end{algorithmic}
\end{algorithm}

\section{Experiments}

\subsection{Implementation Details}
All experiments were conducted on a server equipped with an AMD EPYC 7C13 64-Core Processor and a single NVIDIA L20 GPU. The software environment was built on PyTorch 2.5.1 with CUDA 12.4. For training, we employed a batch size of 16 and an initial learning rate of 2e-5. The initial learning session was trained for 60 epochs, while subsequent incremental sessions were trained for 10 epochs each.

\subsection{Performance of Base Models}
\label{sec:performance_base}

To validate the effectiveness of our selected base models, we evaluated their performance on the CASME II and CAS(ME)$^3$ datasets under the standard Leave-One-Subject-Out (LOSO) protocol. The experiments for 5-class classification on CASME II (Table~\ref{tab:casme2_performance}) and 7-class classification on CAS(ME)$^3$ (Table~\ref{tab:casme3_performance}) were evaluated using accuracy (Acc), F1-score (F1), and Unweighted Average Recall (UAR) metrics. \ZQ{Given the inherent class imbalance in ME datasets, F1 and UAR provide a more comprehensive assessment than accuracy alone. Among the backbones,} our Swin Transformer-based approach achieved 0.815 accuracy and 0.805 F1-score on CASME II, demonstrating competitive performance despite using a general-purpose architecture. \ZQ{Consistent with broader computer vision trends, Transformer-based architectures (ViT and Swin Transformer) demonstrate superior performance, likely benefiting from their ability to capture global dependencies. In contrast, the ResNet backbone exhibits relatively lower performance, potentially challenged by the severe class imbalance and limited sample size inherent in ME tasks.} These results establish a solid baseline foundation for the subsequent incremental learning experiments.

\begin{table}[!t]
\centering
\caption{Performance on CASME II (5-class classification).}
\label{tab:casme2_performance}
\begin{tabular}{lccc}
    \toprule
    \textbf{Model} & \textbf{Acc} & \textbf{F1} & \textbf{UAR} \\
    \midrule
    SLSTT-LSTM~\cite{zhang2022short}       & 0.758 & 0.753 & - \\
    Micron-BERT~\cite{nguyen2023micron}    & -     & 0.855 & 0.835 \\
    Mimicking~\cite{ruan2022mimicking}     & 0.833 & 0.827 & - \\
    MiMaNet~\cite{xia2021micro}          & -     & 0.759 & 0.799 \\
    Swin Transformer                     & 0.815 & 0.805 & 0.793 \\
    ViT                                  & 0.766 & 0.723 & 0.719 \\
    ResNet                               & 0.556 & 0.416 & 0.406 \\
    \bottomrule
\end{tabular}
\end{table}

\begin{table}[!t]
\centering
\caption{Performance on CAS(ME)$^3$ (7-class classification).}
\label{tab:casme3_performance}
\begin{tabular}{lccc}
    \toprule
    \textbf{Model} & \textbf{Acc} & \textbf{F1} & \textbf{UAR} \\
    \midrule
    STSTNet~\cite{liong2019shallow}      & -     & 0.380 & 0.379 \\
    RCN~\cite{xia2020revealing}          & -     & 0.393 & 0.389 \\
    FeatRef~\cite{zhou2022feature}       & -     & 0.349 & 0.341 \\
    Micron-BERT~\cite{nguyen2023micron}    & -     & 0.325 & 0.326 \\
    Swin Transformer                     & 0.609 & 0.465 & 0.454 \\
    ViT                                  & 0.592 & 0.429 & 0.425 \\
    ResNet                               & 0.567 & 0.286 & 0.344 \\
    \bottomrule
\end{tabular}
\end{table}

\subsection{Main Results}
\label{sec:main_results}

\begin{table*}[p]
\centering
\scriptsize
\setlength{\tabcolsep}{3pt}
\renewcommand{\arraystretch}{1.15}
\caption{Experimental Results Across Sessions (ILCV). The columns ``Session 1'' through ``Session 5'' report the Accuracy (\%) at each session. \ZQ{Note that ``Session'' refers to an incremental learning stage where the model adapts to a new dataset.} $\bar{A}$ and $\tilde{A}$ denote Average and Final Accuracy, while \ZQ{$\bar{U}$ and $\bar{F}_1$ represent Average UAR and Average F1.}}
\label{tab:ilcv_results_dfme}
\resizebox{\textwidth}{!}{%
\begin{tabular}{lccccccccc}
    \toprule
    \textbf{Methods} & Session 1 & Session 2 & Session 3 & Session 4 & Session 5 & \textbf{$\bar{A} \uparrow$} & \textbf{$\tilde{A} \uparrow$} & \ZQ{\textbf{$\bar{U} \uparrow$}} & \ZQ{\textbf{$\bar{F}_1 \uparrow$}} \\
    \midrule
    \multicolumn{10}{c}{\textbf{ResNet}} \\
    \midrule
    DER & 38.31 \tiny{$\pm$0.62} & 19.76 \tiny{$\pm$1.46} & 22.96 \tiny{$\pm$1.34} & 26.36 \tiny{$\pm$0.21} & 32.41 \tiny{$\pm$1.33} & 27.96 \tiny{$\pm$0.31} & 32.41 \tiny{$\pm$1.33} & 14.25 \tiny{$\pm$0.34} & 9.70 \tiny{$\pm$0.49} \\
    Finetune & 38.71 \tiny{$\pm$0.62} & 31.07 \tiny{$\pm$0.75} & 26.52 \tiny{$\pm$0.87} & 22.28 \tiny{$\pm$0.60} & 19.84 \tiny{$\pm$0.26} & 27.68 \tiny{$\pm$0.32} & 19.84 \tiny{$\pm$0.26} & 13.42 \tiny{$\pm$0.15} & 8.47 \tiny{$\pm$0.48} \\
    FOSTER & 39.65 \tiny{$\pm$0.13} & 15.93 \tiny{$\pm$0.40} & 24.15 \tiny{$\pm$1.75} & 28.25 \tiny{$\pm$0.21} & 34.36 \tiny{$\pm$0.65} & 28.47 \tiny{$\pm$0.56} & 34.36 \tiny{$\pm$0.65} & 15.95 \tiny{$\pm$0.25} & 11.25 \tiny{$\pm$0.31} \\
    RanPAC & 43.01 \tiny{$\pm$0.59} & 35.60 \tiny{$\pm$1.52} & 34.37 \tiny{$\pm$0.23} & 33.88 \tiny{$\pm$0.45} & 36.11 \tiny{$\pm$0.23} & 36.59 \tiny{$\pm$0.47} & 36.11 \tiny{$\pm$0.23} & 24.37 \tiny{$\pm$0.39} & 23.78 \tiny{$\pm$0.47} \\
    \rowcolor{gray!20}
    Ours & 47.31 \tiny{$\pm$0.97} & 39.51 \tiny{$\pm$1.61} & 37.73 \tiny{$\pm$1.16} & 36.49 \tiny{$\pm$0.13} & 37.00 \tiny{$\pm$0.22} & \textbf{39.61 \tiny{$\pm$0.64}} & \textbf{37.00 \tiny{$\pm$0.22}} & \textbf{28.83 \tiny{$\pm$0.59}} & \textbf{29.14 \tiny{$\pm$0.79}} \\
    \multicolumn{10}{c}{\textbf{ViT}} \\
    \midrule
    DER & 58.87 \tiny{$\pm$1.42} & 20.28 \tiny{$\pm$1.09} & 40.84 \tiny{$\pm$0.87} & 37.42 \tiny{$\pm$0.86} & 33.11 \tiny{$\pm$0.30} & 38.10 \tiny{$\pm$0.32} & 33.11 \tiny{$\pm$0.30} & 27.68 \tiny{$\pm$0.22} & 26.23 \tiny{$\pm$0.24} \\
    DualPrompt & 59.01 \tiny{$\pm$1.32} & 39.43 \tiny{$\pm$1.53} & 44.15 \tiny{$\pm$0.26} & 35.11 \tiny{$\pm$0.26} & 29.99 \tiny{$\pm$0.17} & 41.54 \tiny{$\pm$0.56} & 29.99 \tiny{$\pm$0.17} & 31.76 \tiny{$\pm$0.36} & 31.73 \tiny{$\pm$0.40} \\
    Finetune & 56.18 \tiny{$\pm$0.48} & 38.12 \tiny{$\pm$0.54} & 35.90 \tiny{$\pm$1.43} & 29.36 \tiny{$\pm$0.24} & 23.99 \tiny{$\pm$0.65} & 36.71 \tiny{$\pm$0.60} & 23.99 \tiny{$\pm$0.65} & 28.20 \tiny{$\pm$0.46} & 27.55 \tiny{$\pm$0.34} \\
    FOSTER & 57.53 \tiny{$\pm$1.51} & 24.63 \tiny{$\pm$0.71} & 34.72 \tiny{$\pm$0.87} & 29.33 \tiny{$\pm$1.75} & 26.94 \tiny{$\pm$1.22} & 34.63 \tiny{$\pm$0.68} & 26.94 \tiny[$\pm$1.22] & 26.01 \tiny{$\pm$0.64} & 25.16 \tiny{$\pm$0.79} \\
    L2P & 55.51 \tiny{$\pm$1.75} & 34.90 \tiny{$\pm$2.58} & 39.21 \tiny{$\pm$1.17} & 34.53 \tiny{$\pm$1.11} & 28.30 \tiny{$\pm$0.67} & 38.49 \tiny{$\pm$1.40} & 28.30 \tiny{$\pm$0.67} & 29.59 \tiny{$\pm$0.85} & 29.04 \tiny{$\pm$0.83} \\
    RanPAC & 54.30 \tiny{$\pm$0.97} & 44.91 \tiny{$\pm$0.30} & 50.42 \tiny{$\pm$0.32} & 42.96 \tiny{$\pm$0.26} & 42.45 \tiny{$\pm$0.03} & 47.01 \tiny{$\pm$0.12} & 42.45 \tiny{$\pm$0.03} & 37.86 \tiny{$\pm$0.20} & 37.51 \tiny{$\pm$0.25} \\
    \rowcolor{gray!20}
    Ours & 59.54 \tiny{$\pm$1.15} & 49.26 \tiny{$\pm$1.11} & 51.06 \tiny{$\pm$0.62} & 43.11 \tiny{$\pm$0.39} & 42.88 \tiny{$\pm$0.48} & \textbf{49.17 \tiny{$\pm$0.53}} & \textbf{42.88 \tiny{$\pm$0.48}} & \textbf{40.67 \tiny{$\pm$0.81}} & \textbf{39.64 \tiny{$\pm$0.76}} \\
    \multicolumn{10}{c}{\textbf{SwinT}} \\
    \midrule
    DER & 53.36 \tiny{$\pm$0.82} & 18.89 \tiny{$\pm$0.88} & 36.44 \tiny{$\pm$1.60} & 39.01 \tiny{$\pm$0.26} & 33.26 \tiny{$\pm$0.33} & 36.19 \tiny{$\pm$0.49} & 33.26 \tiny{$\pm$0.33} & 24.09 \tiny{$\pm$0.26} & 21.59 \tiny{$\pm$0.26} \\
    Finetune & 51.48 \tiny{$\pm$2.49} & 38.38 \tiny{$\pm$1.89} & 34.67 \tiny{$\pm$0.52} & 30.30 \tiny{$\pm$0.36} & 25.74 \tiny{$\pm$0.19} & 36.11 \tiny{$\pm$0.93} & 25.74 \tiny{$\pm$0.19} & 25.63 \tiny{$\pm$0.44} & 25.37 \tiny{$\pm$0.67} \\
    FOSTER & 51.75 \tiny{$\pm$0.54} & 26.46 \tiny{$\pm$1.40} & 37.23 \tiny{$\pm$1.10} & 35.11 \tiny{$\pm$0.87} & 28.44 \tiny{$\pm$1.57} & 35.80 \tiny{$\pm$0.86} & 28.44 \tiny{$\pm$1.57} & 25.07 \tiny{$\pm$0.53} & 24.11 \tiny{$\pm$0.50} \\
    RanPAC & 49.06 \tiny{$\pm$1.75} & 43.43 \tiny{$\pm$0.83} & 46.47 \tiny{$\pm$0.13} & 40.24 \tiny{$\pm$0.59} & 42.14 \tiny{$\pm$0.57} & 44.27 \tiny{$\pm$0.26} & 42.14 \tiny{$\pm$0.57} & 35.12 \tiny{$\pm$0.89} & 34.78 \tiny{$\pm$0.79} \\
    \rowcolor{gray!20}
    Ours & 57.12 \tiny{$\pm$0.48} & 49.78 \tiny{$\pm$0.31} & 51.60 \tiny{$\pm$0.05} & 44.51 \tiny{$\pm$0.10} & 44.35 \tiny{$\pm$0.08} & \textbf{49.48 \tiny{$\pm$0.05}} & \textbf{44.35 \tiny{$\pm$0.08}} & \textbf{40.07 \tiny{$\pm$0.77}} & \textbf{39.87 \tiny{$\pm$0.46}} \\
    \bottomrule
\end{tabular}
}%

\vspace{0.5cm}

\caption{Experimental Results Across Sessions (SLCV). The columns ``Session 1'' through ``Session 5'' report the Accuracy (\%) at each session. $\bar{A}$ and $\tilde{A}$ denote Average and Final Accuracy, while \ZQ{$\bar{U}$ and $\bar{F}_1$ represent Average UAR and Average F1.}}
\label{tab:slcv_results_dfme}
\resizebox{\textwidth}{!}{%
\begin{tabular}{lccccccccc}
    \toprule
    \textbf{Methods} & Session 1 & Session 2 & Session 3 & Session 4 & Session 5 & \textbf{$\bar{A} \uparrow$} & \textbf{$\tilde{A} \uparrow$} & \ZQ{\textbf{$\bar{U} \uparrow$}} & \ZQ{\textbf{$\bar{F}_1 \uparrow$}} \\
    \midrule
    \multicolumn{10}{c}{\textbf{ResNet}} \\
    \midrule
    DER & 36.96 \tiny{$\pm$0.59} & 21.15 \tiny{$\pm$1.57} & 23.46 \tiny{$\pm$1.83} & 26.09 \tiny{$\pm$0.37} & 32.17 \tiny{$\pm$1.05} & 27.97 \tiny{$\pm$0.35} & 32.17 \tiny{$\pm$1.05} & 14.28 \tiny{$\pm$0.33} & 10.06 \tiny{$\pm$0.24} \\
    Finetune & 37.37 \tiny{$\pm$1.64} & 29.07 \tiny{$\pm$1.45} & 26.27 \tiny{$\pm$0.84} & 21.71 \tiny{$\pm$0.40} & 21.43 \tiny{$\pm$1.18} & 27.17 \tiny{$\pm$0.59} & 21.43 \tiny{$\pm$1.18} & 13.23 \tiny{$\pm$0.12} & 8.59 \tiny{$\pm$0.26} \\
    FOSTER & 37.37 \tiny{$\pm$1.78} & 17.15 \tiny{$\pm$1.52} & 24.25 \tiny{$\pm$1.14} & 28.66 \tiny{$\pm$0.52} & 34.84 \tiny{$\pm$1.01} & 28.45 \tiny{$\pm$0.62} & 34.84 \tiny{$\pm$1.01} & 15.75 \tiny{$\pm$0.21} & 11.19 \tiny{$\pm$0.29} \\
    RanPAC & 34.01 \tiny{$\pm$1.75} & 29.77 \tiny{$\pm$0.69} & 33.78 \tiny{$\pm$0.56} & 30.76 \tiny{$\pm$0.77} & 33.55 \tiny{$\pm$0.30} & 32.37 \tiny{$\pm$0.26} & 33.55 \tiny{$\pm$0.30} & 21.06 \tiny{$\pm$0.60} & 20.35 \tiny{$\pm$0.81} \\
    \rowcolor{gray!20}
    Ours & 38.98 \tiny{$\pm$0.36} & 35.25 \tiny{$\pm$0.66} & 33.73 \tiny{$\pm$0.13} & 31.62 \tiny{$\pm$0.44} & 34.17 \tiny{$\pm$0.61} & \textbf{34.75 \tiny{$\pm$0.15}} & \textbf{34.17 \tiny{$\pm$0.61}} & \textbf{23.94 \tiny{$\pm$0.96}} & \textbf{23.50 \tiny{$\pm$1.27}} \\
    \multicolumn{10}{c}{\textbf{ViT}} \\
    \midrule
    DER & 47.85 \tiny{$\pm$1.42} & 20.10 \tiny{$\pm$0.84} & 40.15 \tiny{$\pm$0.97} & 35.45 \tiny{$\pm$0.56} & 32.53 \tiny{$\pm$0.28} & 35.22 \tiny{$\pm$0.57} & 32.53 \tiny{$\pm$0.28} & 24.84 \tiny{$\pm$0.48} & 23.41 \tiny{$\pm$0.42} \\
    DualPrompt & 46.64 \tiny{$\pm$1.36} & 31.77 \tiny{$\pm$1.26} & 39.06 \tiny{$\pm$0.05} & 32.60 \tiny{$\pm$0.22} & 29.26 \tiny{$\pm$0.59} & 35.87 \tiny{$\pm$0.58} & 29.26 \tiny{$\pm$0.59} & 25.96 \tiny{$\pm$0.29} & 25.64 \tiny{$\pm$0.29} \\
    Finetune & 48.79 \tiny{$\pm$2.03} & 36.29 \tiny{$\pm$1.74} & 34.91 \tiny{$\pm$0.27} & 28.35 \tiny{$\pm$0.58} & 23.53 \tiny{$\pm$0.85} & 34.38 \tiny{$\pm$0.58} & 23.53 \tiny{$\pm$0.85} & 25.14 \tiny{$\pm$0.25} & 24.76 \tiny{$\pm$0.17} \\
    FOSTER & 47.31 \tiny{$\pm$0.88} & 23.15 \tiny{$\pm$0.86} & 30.02 \tiny{$\pm$1.97} & 29.04 \tiny{$\pm$1.14} & 22.83 \tiny{$\pm$1.75} & 30.47 \tiny{$\pm$0.96} & 22.83 \tiny{$\pm$1.75} & 22.65 \tiny{$\pm$0.37} & 21.30 \tiny{$\pm$0.49} \\
    L2P & 49.60 \tiny{$\pm$1.01} & 29.16 \tiny{$\pm$0.61} & 36.15 \tiny{$\pm$0.65} & 31.39 \tiny{$\pm$1.26} & 26.89 \tiny{$\pm$0.46} & 34.63 \tiny{$\pm$0.71} & 26.89 \tiny{$\pm$0.46} & 25.83 \tiny{$\pm$0.25} & 25.53 \tiny{$\pm$0.29} \\
    RanPAC & 47.45 \tiny{$\pm$0.97} & 39.43 \tiny{$\pm$0.66} & 42.91 \tiny{$\pm$0.39} & 38.09 \tiny{$\pm$0.24} & 39.97 \tiny{$\pm$0.13} & 41.57 \tiny{$\pm$0.14} & 39.97 \tiny{$\pm$0.13} & 31.53 \tiny{$\pm$0.26} & 30.81 \tiny{$\pm$0.25} \\
    \rowcolor{gray!20}
    Ours & 48.52 \tiny{$\pm$0.82} & 42.12 \tiny{$\pm$0.57} & 45.19 \tiny{$\pm$0.23} & 39.72 \tiny{$\pm$0.63} & 40.39 \tiny{$\pm$0.36} & \textbf{43.19 \tiny{$\pm$0.43}} & \textbf{40.39 \tiny{$\pm$0.36}} & \textbf{34.60 \tiny{$\pm$0.51}} & \textbf{33.87 \tiny{$\pm$0.31}} \\
    \multicolumn{10}{c}{\textbf{SwinT}} \\
    \midrule
    DER & 44.09 \tiny{$\pm$1.17} & 16.45 \tiny{$\pm$1.23} & 36.35 \tiny{$\pm$1.39} & 35.82 \tiny{$\pm$0.34} & 33.52 \tiny{$\pm$0.26} & 33.24 \tiny{$\pm$0.38} & 33.52 \tiny{$\pm$0.26} & 21.36 \tiny{$\pm$0.12} & 18.93 \tiny{$\pm$0.28} \\
    Finetune & 43.28 \tiny{$\pm$0.67} & 32.03 \tiny{$\pm$0.83} & 32.25 \tiny{$\pm$1.19} & 27.70 \tiny{$\pm$0.98} & 23.98 \tiny{$\pm$0.76} & 31.85 \tiny{$\pm$0.51} & 23.98 \tiny{$\pm$0.76} & 21.21 \tiny{$\pm$0.19} & 20.44 \tiny{$\pm$0.09} \\
    FOSTER & 42.74 \tiny{$\pm$2.64} & 21.32 \tiny{$\pm$0.38} & 33.04 \tiny{$\pm$0.78} & 33.69 \tiny{$\pm$0.69} & 29.14 \tiny{$\pm$1.41} & 31.99 \tiny{$\pm$0.50} & 29.14 \tiny{$\pm$1.41} & 21.27 \tiny{$\pm$0.70} & 19.98 \tiny{$\pm$0.55} \\
    RanPAC & 39.25 \tiny{$\pm$0.94} & 37.34 \tiny{$\pm$0.45} & 40.49 \tiny{$\pm$1.02} & 34.61 \tiny{$\pm$0.07} & 38.65 \tiny{$\pm$0.20} & 38.07 \tiny{$\pm$0.10} & 38.65 \tiny{$\pm$0.20} & 27.89 \tiny{$\pm$0.38} & 27.67 \tiny{$\pm$0.32} \\
    \rowcolor{gray!20}
    Ours & 47.98 \tiny{$\pm$1.07} & 41.43 \tiny{$\pm$1.13} & 45.23 \tiny{$\pm$0.69} & 39.99 \tiny{$\pm$0.82} & 41.76 \tiny{$\pm$0.07} & \textbf{43.28 \tiny{$\pm$0.07}} & \textbf{41.76 \tiny{$\pm$0.07}} & \textbf{32.71 \tiny{$\pm$0.56}} & \textbf{32.14 \tiny{$\pm$0.59}} \\
    \bottomrule
\end{tabular}
}%
\end{table*}

\begin{figure*}[!t] 
    \centering
    \includegraphics[width=1.0\linewidth]{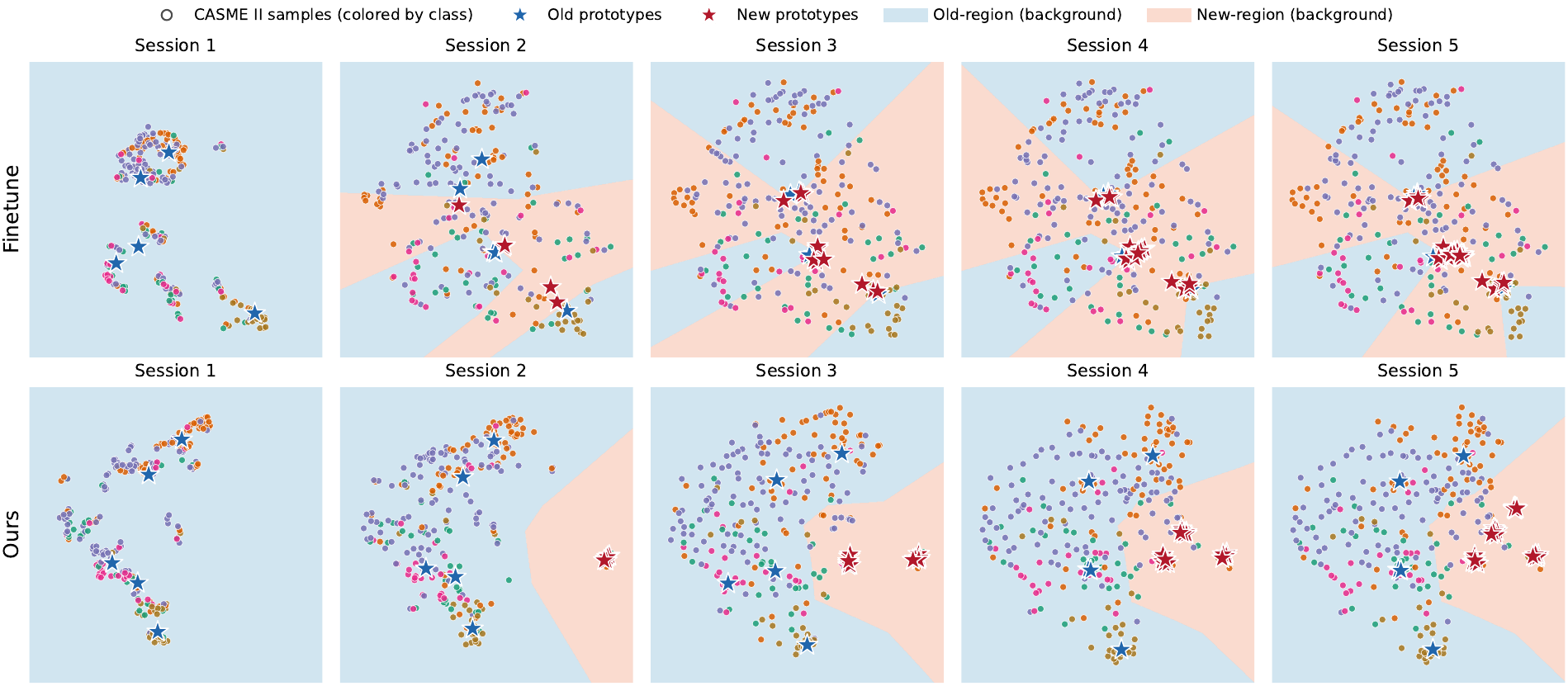}
    \caption{\ZQ{t-SNE visualization of CASME II samples (colored dots) throughout the incremental sessions. Blue stars denote prototypes of CASME II classes (Old), while red stars denote prototypes of new classes added in subsequent sessions (New). The background colors indicate decision regions (Blue: Old-region; Red: New-region). Compared to Finetune (top), our method (bottom) better preserves the structure and decision boundaries for old samples.}}
    \label{fig:domain_shift}
\end{figure*}

\begin{figure}[!t] 
    \centering
    \includegraphics[width=1.0\columnwidth]{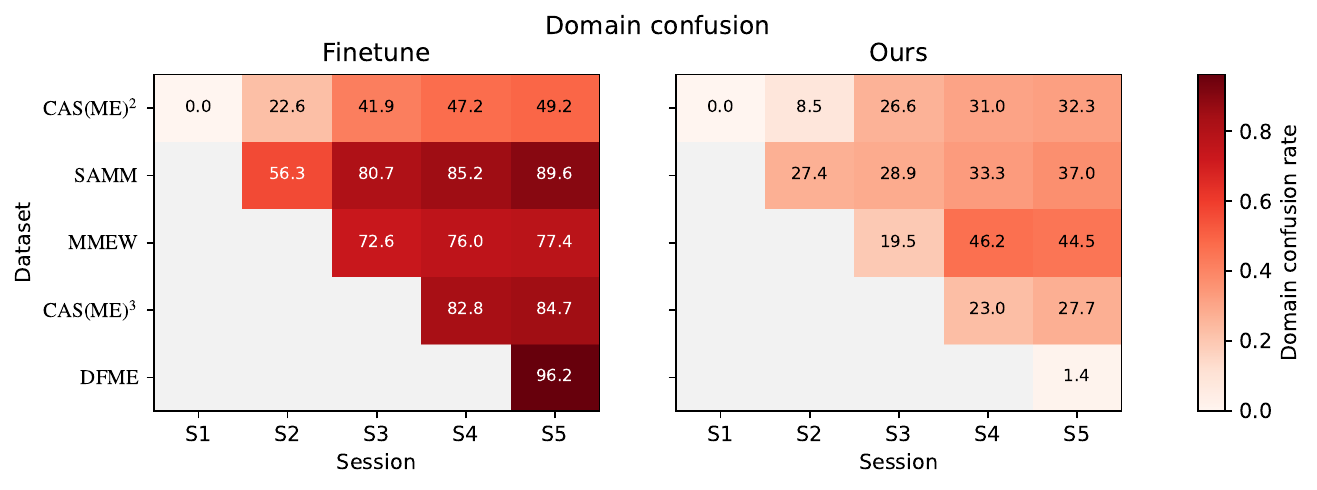}
    \caption{\ZQ{Domain confusion rates across sessions. Each cell reports the percentage of misclassified samples predicted as classes belonging to a different session (domain) rather than their source session. Lower values indicate reduced cross-session interference. Compared to Finetune, our method significantly reduces this confusion, corroborating the t-SNE visualization (Fig. \ref{fig:domain_shift}) and demonstrating its effectiveness in mitigating domain confusion.}}
    \label{fig:domain_confusion}
\end{figure}

We first present the main incremental learning performance across different backbones and testing protocols in Table~\ref{tab:ilcv_results_dfme} and Table~\ref{tab:slcv_results_dfme}. Consistent with findings in conventional MER, transformer-based backbones generally outperform ResNet architectures in the incremental setting.

Further analysis reveals distinct performance trends among different categories of incremental learning methods. Overall, class-prototype-based approaches (RanPAC, Ours) demonstrate the best performance, followed by prompt-based methods (L2P, DualPrompt), while dynamic network methods (FOSTER, DER) exhibit the lowest accuracy. \ZQ{To investigate why our method achieves superior performance, we analyze the extent of \emph{cross-session misclassification} across different learning stages. This metric reflects the tendency of samples being predicted as classes from other sessions, which is a direct consequence of catastrophic forgetting and domain confusion. While the first major drop often occurs at the first dataset switch (e.g., Session 1 $\rightarrow$ Session 2), the more critical observation is that this \emph{domain confusion accumulates over time} as more sessions are introduced.}

\ZQ{Figure~\ref{fig:domain_shift} visualizes this progressive interference on CASME II samples. Under Finetune, prototypes of newly introduced classes repeatedly encroach upon the old feature space and expand the new-region, shrinking the effective decision region for old classes. In contrast, our method better preserves the old prototypes and decision regions across sessions, reducing the tendency to overwrite earlier knowledge.}

\ZQ{This observation is further quantified in Figure~\ref{fig:domain_confusion}, where we compare the prediction distribution of misclassified samples. Using Finetune as a representative baseline, we find that a large fraction of errors are wrongly predicted as classes belonging to a different session (domain). For example, when SAMM is first introduced (Session~2), 56.3\% of its misclassifications are already mapped to the previous session. By Session~5, the confusion rate reaches 49.2\% for CASME II (older $\rightarrow$ newer) and 96.2\% for DFME (newer $\rightarrow$ older). Conversely, our method significantly lowers this cross-session confusion (e.g., 27.4\% for SAMM at Session~2, 32.3\% for CASME II at Session~5, and 1.4\% for DFME at Session~5). This indicates that our method effectively maintains category space stability, ensuring that even when mistakes occur, predictions are more likely to stay within the correct domain. This capability to mitigate domain confusion explains the robustness and effectiveness of our approach.}

The main results presented herein establish the first comprehensive baseline for IMER across various backbones and incremental learning strategies. This benchmark provides valuable insights into the challenges and potential solutions for continually adapting MER systems to new data distributions.

\subsection{Ablation Study}
\label{sec:ablation}
To dissect the contribution of key components of our method, 
we conducted an ablation study using the SwinT backbone \ZQ{under both SLCV and ILCV protocols}, as reported in Table~\ref{tab:ablation_dfme}. \ZQ{While SLCV serves as our primary protocol for assessing generalization, the conclusions regarding component contributions remain consistent across both protocols.} Detailed discussions are provided in the following paragraphs. To establish a fundamental baseline, we include a Nearest Class Mean (NCM) model, where prototypes are simply the mean features of each class. This helps quantify the advantage of our more sophisticated statistical modeling in GEM.

\begin{table*}[!t]
\centering
\caption{Ablation study of different prototype strategies and refinement components, using SwinT under the SLCV and ILCV protocols. $L_a$ denotes ArcFace loss and $L_m$ denotes Mahalanobis regularization. Our full method is highlighted in gray.}
\label{tab:ablation_dfme}
\begin{tabular}{llc cccc|cccc}
\toprule
\multicolumn{2}{l}{Prototype Strategy} & Refinement & \multicolumn{4}{c|}{SLCV} & \multicolumn{4}{c}{ILCV} \\
 & & & \textbf{$\bar{A}$} & \textbf{$\tilde{A}$} & \ZQ{\textbf{$\bar{U}$}} & \ZQ{\textbf{$\bar{F}_1$}} & \textbf{$\bar{A}$} & \textbf{$\tilde{A}$} & \ZQ{\textbf{$\bar{U}$}} & \ZQ{\textbf{$\bar{F}_1$}} \\
\midrule
\rowcolor{gray!20}
GEM (Ours) & (Eq.~\ref{eq:final_wt}) & $L_a+L_m$ & 43.28 & 41.76 & 32.71 & 32.14 & 49.48 & 44.35 & 40.07 & 39.87 \\
GEM & (Eq.~\ref{eq:final_wt}) & $L_a$ & 42.18 & 41.33 & 31.38 & 30.86 & 48.73 & 44.08 & 39.73 & 39.38 \\
GEM & (Eq.~\ref{eq:final_wt}) & $\times$ & 39.68 & 39.70 & 31.25 & 30.56 & 47.49 & 43.14 & 39.65 & 38.88 \\
\midrule
w/o Accum. H & (Eq.~\ref{eq:accumulated_m_wt}) & $L_a+L_m$ & 39.11 & 34.89 & 27.43 & 26.71 & 46.69 & 36.40 & 34.12 & 33.60 \\
w/o Accum. H & (Eq.~\ref{eq:accumulated_m_wt}) & $\times$ & 38.36 & 31.70 & 26.17 & 25.18 & 44.43 & 35.82 & 33.57 & 32.34 \\
\midrule
w/o Accum. M \& H & (Eq.~\ref{eq:separate_wt}) & $L_a+L_m$ & 37.65 & 31.87 & 26.56 & 25.96 & 44.69 & 34.05 & 33.26 & 32.19 \\
w/o Accum. M \& H & (Eq.~\ref{eq:separate_wt}) & $\times$ & 35.97 & 30.77 & 24.89 & 24.19 & 41.81 & 33.03 & 32.48 & 31.75\\
\midrule
NCM (Baseline) & & $\times$ & 29.26 & 26.02 & 24.34 & 21.59 & 35.92 & 29.55 & 31.86 & 25.61 \\
\bottomrule
\end{tabular}
\end{table*}

\subsubsection{Importance of General Evolution via Statistic Accumulation}

In Table~\ref{tab:ablation_dfme}, we evaluate different prototype strategies. Our full method, using GEM with full refinement, achieves the best final accuracy under both protocols: $\tilde{A}=\ZQ{41.76\%}$ (SLCV) and $\tilde{A}=\ZQ{44.35\%}$ (ILCV) (\emph{row 1}).

\begin{figure}[!t]
\centering
\includegraphics[width=1\columnwidth]{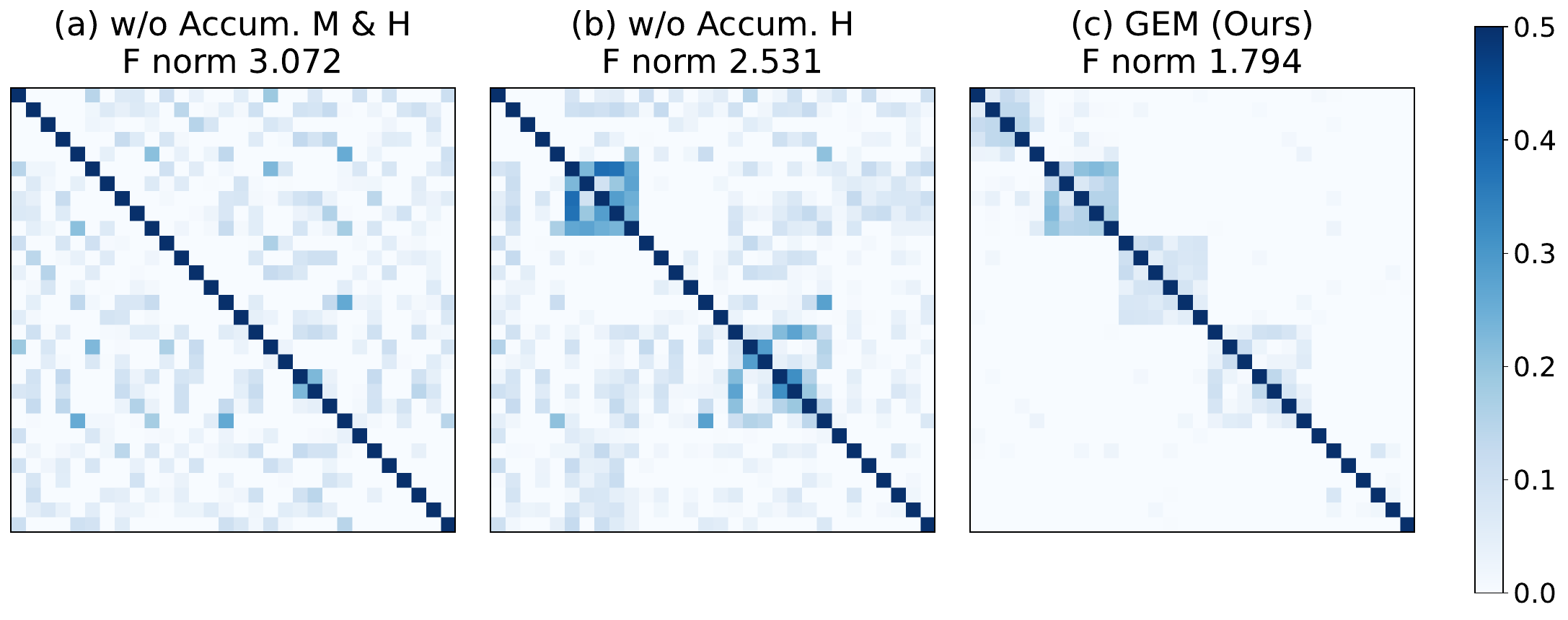}
\caption{{Correlation coefficient matrices between class prototypes across all sessions, using different prototype update strategies. (a) Using separate $M^t$ matrices per session (Eq.~\ref{eq:separate_wt}). (b) Accumulating only the $M$ matrix (Eq.~\ref{eq:accumulated_m_wt}). (c) Our GEM approach (Eq.~\ref{eq:final_wt}), which accumulates both $M$ and $H$ matrices. Lower Frobenius norm indicates better overall orthogonality.}}

\label{fig:pearson}
\end{figure}

In contrast, if we use a strategy \emph{w/o Accum. M \& H}, where statistics are not accumulated and weights are computed independently for each session:
\begin{equation}
W^{t}_a = (M^t)^{-1} H^t,
\label{eq:separate_wt} 
\end{equation}
where $M^t = (X^t)^TX^t+\lambda I$ and $H^t = (X^t)^TY^t$, the performance with full refinement drops to $\tilde{A}=\ZQ{31.87\%}$ (SLCV) and $\tilde{A}=\ZQ{34.05\%}$ (ILCV) (\emph{row 6}). This comparison highlights that using accumulated statistics substantially outperforms using fragmented, session-specific ones. Fig.~\ref{fig:pearson}~(a) shows that the \emph{w/o Accum. M \& H} strategy leads to high inter-prototype correlations \ZQ{(Frobenius norm = 3.072)}, whereas our GEM approach significantly reduces these correlations \ZQ{(Frobenius norm = 1.794)}, as shown in Fig.~\ref{fig:pearson}~(c).

\subsubsection{Benefit of Evolving All Class Prototypes}
We further contrast our GEM approach (accumulating both $M$ and $H$) with an alternative that proceeds \emph{w/o Accum. H} (i.e., only accumulating the second-order statistics), computing prototypes as:
\begin{equation}
W^{t}_b = (M_c^{t})^{-1} H^t.
\label{eq:accumulated_m_wt}
\end{equation}
This latter approach, which effectively freezes prior prototypes relative to the evolving metric $(M_c^{t})^{-1}$, yields a much lower final accuracy of $\tilde{A}=\ZQ{34.89\%}$ (SLCV) and $\tilde{A}=\ZQ{36.40\%}$ (ILCV) (\emph{row 4}), compared to our \ZQ{41.76\%}/\ZQ{44.35\%} (SLCV/ILCV) (\emph{row 1}). This demonstrates the significant advantage of our holistic update, which allows joint optimization of all prototypes by accumulating $H_c^t$. The superior inter-prototype orthogonality of our approach is also evident in Fig.~\ref{fig:pearson}c compared to Fig.~\ref{fig:pearson}b.

\subsubsection{Contribution of Local Adaptation Components (\texorpdfstring{$L_a$ and $L_m$}{La and Lm})}
Finally, we evaluate the impact of the local adaptation components, $L_a$ (ArcFace loss) and $L_m$ (Mahalanobis regularization), applied to the prototypes from GEM. Without any refinement, the GEM strategy achieves $\tilde{A}=\ZQ{39.70\%}$ (SLCV) and $\tilde{A}=\ZQ{43.14\%}$ (ILCV) (\emph{row 3}). Adding only the ArcFace refinement ($L_a$) increases this to \ZQ{41.33\%}/\ZQ{44.08\%} (SLCV/ILCV) (\emph{row 2}), demonstrating the benefit of enforcing larger class margins. Incorporating both $L_a$ and $L_m$ constitutes our full method, reaching \ZQ{41.76\%}/\ZQ{44.35\%} (SLCV/ILCV) (\emph{row 1}). Although the numerical improvement over using $L_a$ alone may appear modest, the primary advantage of $L_m$ lies in its ability to enforce stability during the refinement process, as will be further detailed in the \textbf{Mahalanobis Constraint for Balanced Refinement} analysis (Section~\ref{sec:AMR}). The consistent performance gains from refinement are observed across different prototype strategies (e.g., comparing \emph{rows 4 vs. 5} and \emph{rows 6 vs. 7}), confirming the robustness of our local adaptation module.

\subsection{Computational Cost Analysis}
We analyze the scalability of our method by examining its parameter and time costs. The training times reported in Table~\ref{tab:cost_comparison} represent the total duration for one complete incremental learning fold under the ILCV protocol with the ViT backbone, measured in seconds on our experimental platform.

\begin{table}[!t]
\centering
\caption{Comparison of total parameters and training time for different methods with the ViT-B/16 backbone.}
\label{tab:cost_comparison}
\begin{tabular}{lrc}
\toprule
\textbf{Method} & \textbf{Parameters} & \textbf{Training Time (s)} \\
\midrule
Ours        & 86,411,520  & 169.38 \\
DER         & 429,114,662 & 715.92 \\
L2P         & 171,708,732 & 2006.56 \\
DualPrompt  & 171,896,892 & 1934.13 \\
FOSTER      & 257,528,975 & 815.84 \\
Finetune    & 85,821,726  & 291.98 \\
RanPAC      & 86,098,656  & 365.28 \\
\bottomrule
\end{tabular}
\end{table}

\begin{figure}[!t]
    \centering
    \includegraphics[width=\linewidth]{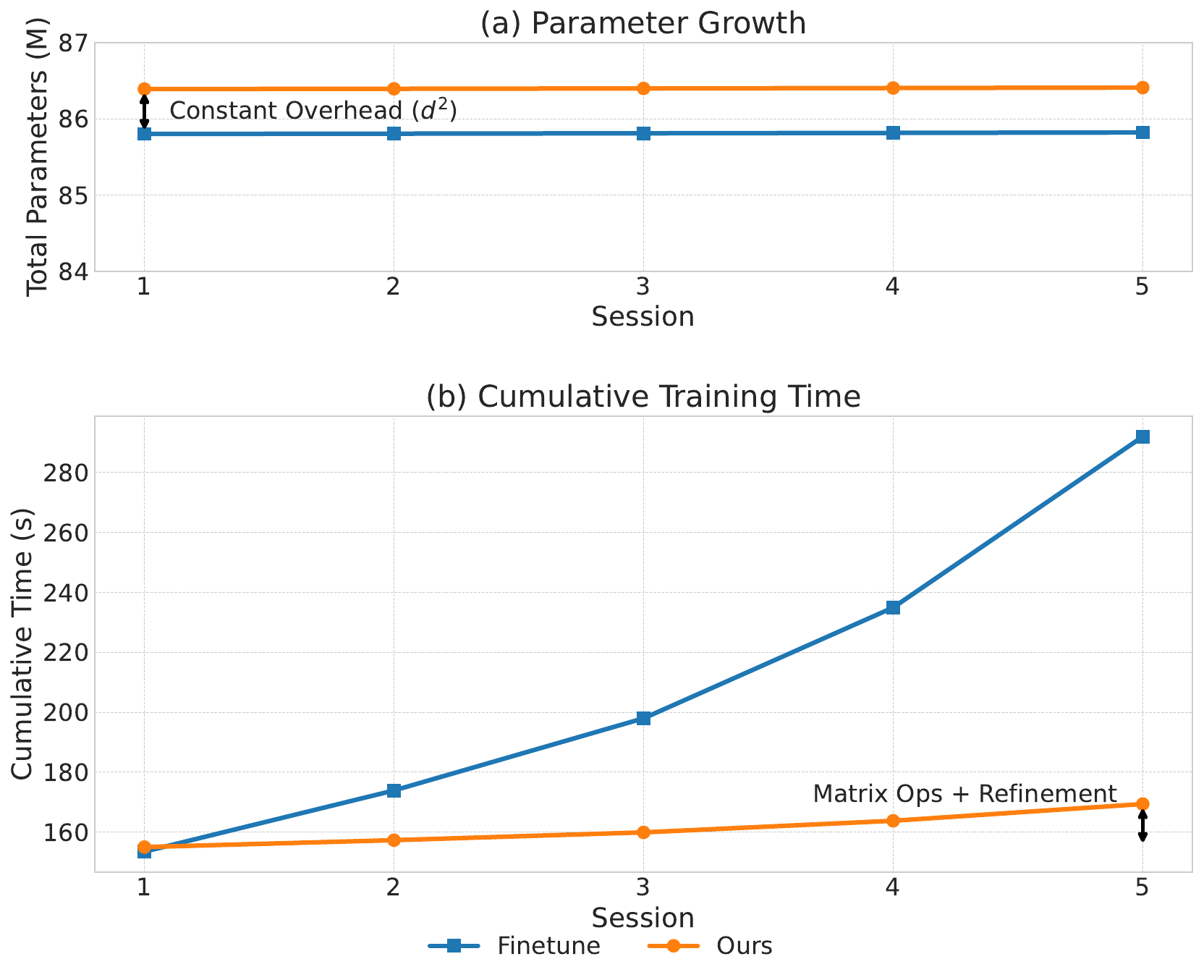}
    \caption{Comparison of computational cost growth between our method and Finetune. (a) Parameter Growth: Compared to Finetune, our method introduces an additional small, constant overhead ($d^2$) for storing the second-order matrix $M$. (b) Training Time Growth: Our method's training time remains minimal in subsequent sessions as it only refines the classifier, whereas Finetune requires more time.}
    \label{fig:cost_analysis}
\end{figure}

\textbf{Space Cost.} The space cost of our method consists of three parts: the backbone parameters ($P_B$), the second-order matrix $M$ ($d \times d$), and the first-order matrix $H$ ($d \times K_t$, where $K_t$ is the number of classes at session $t$). In contrast, the Finetune baseline consists of the backbone and a classification head ($(d+1) \times K_t$). As illustrated in Figure~\ref{fig:cost_analysis}(a), the primary difference is the $O(d^2)$ term for matrix $M$ in our method, which is constant and does not grow with sessions. For a ViT backbone with $d=768$, this $d^2$ term constitutes only about 0.69\% of the total model parameters, representing a minimal and fixed overhead.

\textbf{Time Cost.} The time cost of our method includes initial backbone fine-tuning, per-session matrix operations, and classifier refinement. As shown in Figure~\ref{fig:cost_analysis}(b), after the initial session, our method's incremental training time (for sessions $>1$) is significantly lower than Finetune. This is because our approach only requires lightweight classifier refinement and matrix operations in subsequent sessions, avoiding the need to retrain the entire backbone. The cost of per-session matrix operations is constant and negligible (approx. 0.02s for a $d=768$ matrix inversion on a standard CPU). This efficiency, also reflected in Table~\ref{tab:cost_comparison}, makes our method highly practical for real-world incremental learning scenarios.

Overall, our method demonstrates significant advantages in both space and time scalability, making it highly practical for real-world incremental learning scenarios.

\subsection{Hyperparameter Selection}
\label{sec:appendix_sensitivity_alpha}

\begin{figure}[!t]
\centering
\includegraphics[width=0.8\columnwidth]{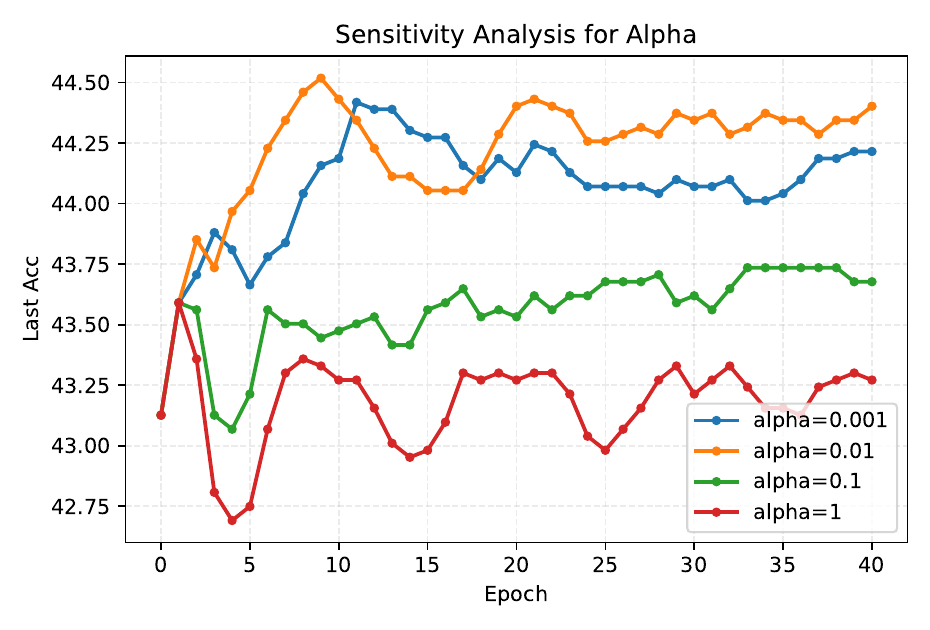} 
\caption{Sensitivity analysis of the hyperparameter $\alpha$. The plot shows the final session accuracy ($\tilde{A}$) on the validation set over training epochs for different values of $\alpha$.}
\label{fig:sensitivity_alpha}
\end{figure}

The hyperparameter $\alpha$ in Eq.~\ref{eq:combined_loss} is crucial for balancing the discriminative ArcFace loss ($L_a$) and the stabilizing Mahalanobis regularization term ($L_m$). Our sensitivity analysis, depicted in Figure~\ref{fig:sensitivity_alpha}, underscores the necessity of the Mahalanobis constraint. When the constraint is removed ($\alpha=0$), the model's performance becomes highly sensitive to the number of training epochs; it peaks early and then degrades substantially, indicating instability. Conversely, a very strong constraint ($\alpha=1$) ensures stable performance over epochs but limits the model's plasticity, resulting in suboptimal accuracy. A smaller weight ($\alpha=0.001$) proves insufficient to prevent performance decay. Therefore, a well-calibrated trade-off is essential. Based on these results, we selected $\alpha=0.01$ for our experiments.

Another critical hyperparameter is $\lambda$, the L2 regularization coefficient for the moment matrix $M^t = (X^t)^TX^t + \lambda I$. Its role is to ensure $M^t$ is well-conditioned and invertible. Unlike $\alpha$, which is fixed, $\lambda$ is determined for each session by holding out a small portion of its training data as a validation set. This allows us to select the best-performing value from candidates of the form $10^k$.

\section{Discussion}
\subsection{Analysis of the Mahalanobis Refinement Method}
\label{sec:AMR}
The superior performance of our MR approach stems from its two-stage design, addressing both stability and plasticity, centered around Mahalanobis concepts. We analyze these advantages below:

\textbf{Stability through Implicit De-correlation}: The stability primarily stems from the GEM construction defined in Eq.~\ref{eq:final_wt}. This model inherently achieves feature de-correlation, enhancing stability by operating in a feature space endowed with the Mahalanobis metric induced by $(M_c^t)^{-1}$~\cite{de2000mahalanobis,kulis2013metric}.

To understand this, consider the goal of feature whitening~\cite{bishop2006pattern}. If we denote the aggregated features from all sessions up to $t$ as $X_{1..t}$, whitening aims to find a transformation matrix $G$ such that the transformed features have an identity covariance matrix. This implies $G^T G \approx ((X_{1..t})^T X_{1..t})^{-1}$. Our accumulated second-order moment matrix $M_c^t$ approximates the uncentered covariance of all seen data, i.e., $M_c^t \approx (X_{1..t})^T X_{1..t} + \lambda I$. Thus, the whitening transformation requirement becomes $G^T G \approx (M_c^t)^{-1}$. Similarity measurement between a test sample $x_{test}$ and a class representation $x_k$ in this optimally whitened space would be: 
\begin{equation}
    (G x_{test})^T (G x_k) = x_{test}^T G^T G x_k \approx x_{test}^T (M_c^t)^{-1} x_k. 
\end{equation}

This expression shows that comparison in the whitened space corresponds to an inner product under the Mahalanobis metric induced by $(M_c^t)^{-1}$. The associated squared Mahalanobis distance between $x_{test}$ and $x_k$ is $d^2=(x_{test}-x_k)^T (M_c^t)^{-1} (x_{test}-x_k)$, which expands to self-terms minus $2x_{test}^T (M_c^t)^{-1} x_k$; therefore, the similarity term $x_{test}^T (M_c^t)^{-1} x_k$ is directly related to (negative) Mahalanobis distance up to additive and norm terms~\cite{de2000mahalanobis,bishop2006pattern}.

Our GEM achieves this de-correlation implicitly. We do not compute the whitening matrix $G$ explicitly. Instead, the classifier weights themselves are computed as $W^t = (M_c^{t})^{-1} H_c^{t}$. Therefore, the classification score: 
\begin{equation}
x_{test}^T W^t = x_{test}^T (M_c^{t})^{-1} H_c^{t},
\end{equation}
directly implements the desired comparison within this  de-correlated space. By incorporating $(M_c^{t})^{-1}$ into the prototype calculation, the model inherently suppresses feature correlations learned across all sessions. This enhances the orthogonality between class prototypes, significantly mitigating catastrophic forgetting and cross-session interference, particularly when encountering domain shifts, as empirically supported by the stable performance transition from Session 1 to Session 2 and the low inter-prototype correlation shown in Figure~\ref{fig:pearson}(c).

\begin{figure}[!t]
    \centering
    \includegraphics[width=1.0\linewidth]{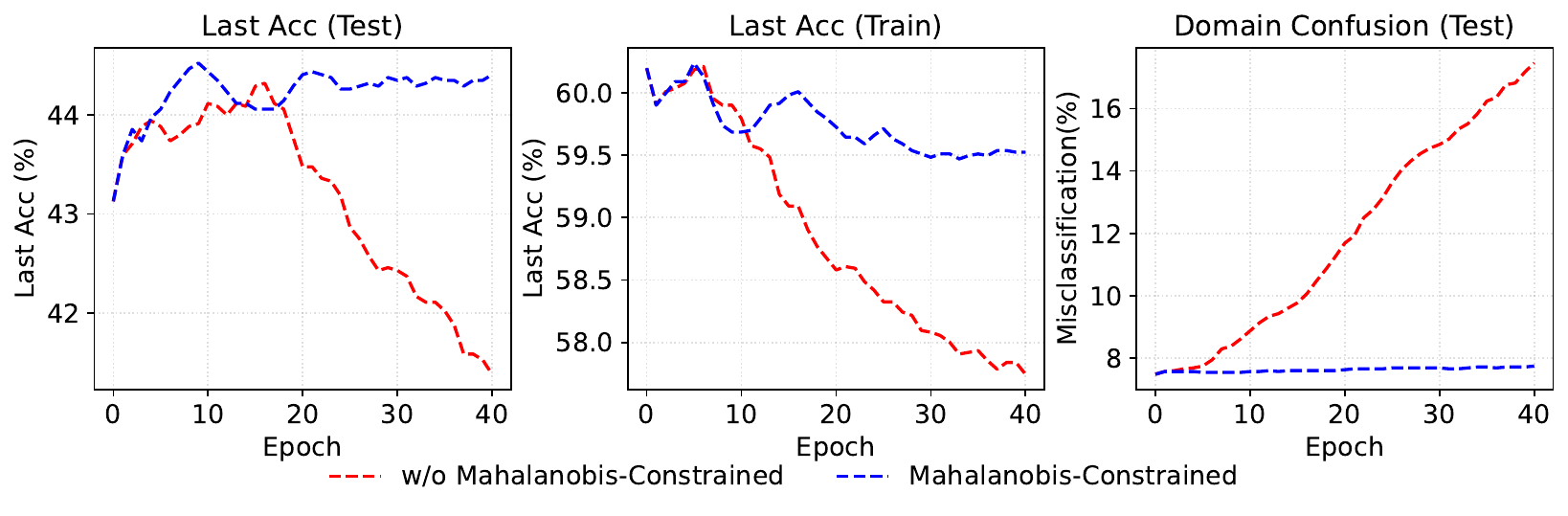}
    \caption{Refinement with (blue) vs. without (red) Mahalanobis constraint. The domain confusion rate is the percentage of samples misclassified into a class belonging to a different session.}
    \label{fig:ablation_plots}
\end{figure}

\textbf{Mahalanobis Constraint for Balanced Refinement}: While GEM ensures stability, refinement is necessary to enhance discriminative power, i.e., plasticity. However, solely refining with a gradient-based loss like ArcFace ($L_a$) proves detrimental. As shown in Figure~\ref{fig:ablation_plots}, unconstrained refinement leads to an initial accuracy increase followed by a rapid performance decline on both training and testing sets. This instability is not conventional overfitting; rather, it arises from disrupting the generally learned prototype structure and orthogonality, resulting in significantly increased domain confusion, where samples are misclassified into incorrect classes from incorrect sessions.

This instability issue demonstrates the necessity of maintaining stability while pursuing plasticity. Our Mahalanobis constraint ($L_m$) balances plasticity and stability by regularizing weight updates. It penalizes deviations from $W_{init}$ in the Mahalanobis space defined by $M_c^t$, anchoring refinement. This preserves general structure (stability) while enabling $L_a$'s discriminative gains (plasticity), preventing instability and boosting final accuracy.

\subsection{Analysis of Intra-Session Correlation}

Figure~\ref{fig:pearson}c, while demonstrating the best overall inter-prototype orthogonality (lowest Frobenius norm) for Eq.~\ref{eq:final_wt}, visually reveals higher correlation coefficients among the prototypes corresponding specifically to Session 2 (SAMM dataset). This manifests as a distinct block of relatively stronger correlation values within the diagonal region associated with Session 2's classes, indicating less effective de-correlation within that specific session compared to the overall set. This occurs because the general implicit whitening via $(M_c^{t})^{-1}$ may compromise optimal de-correlation within a specific session if its data distribution significantly deviates from the cumulative average, prioritizing inter-session orthogonality instead.

Assuming zero-centered Gaussian distributions $P_i = \mathcal{N}(0, \Sigma_i)$ for session $i$ and $P_{all} = \mathcal{N}(0, \Sigma_{all})$ for the accumulated data, the general whitening relates to a transformation $G_{all}$ where $G_{all}^T G_{all} = \Sigma_{all}^{-1} \approx (M_c^t)^{-1}$. Session $i$'s covariance after this general transform is $\Sigma_{actual} = G_{all} \Sigma_i G_{all}^T$. The sub-optimality of this whitening for session $i$, measured by its deviation from an ideal identity covariance $I$, is given by the KL divergence:
\begin{align*}
    \MoveEqLeft[2] \mathbb{D}_{KL}(\mathcal{N}(0, \Sigma_{actual}) || \mathcal{N}(0, I)) \\
    &= \frac{1}{2} \bigl(\text{tr}(\Sigma_i \Sigma_{all}^{-1}) - d - \ln \det(\Sigma_i \Sigma_{all}^{-1})\bigr) \\
    &= \mathbb{D}_{KL}(\mathcal{N}(0, \Sigma_i) || \mathcal{N}(0, \Sigma_{all}))
\end{align*}
Thus, the imperfection of the general whitening for session $i$ equals the KL divergence between its distribution $P_i$ and the overall accumulated distribution $P_{all}$. A larger distributional shift for a session (like Session 2 introducing SAMM) directly corresponds to less effective intra-session de-correlation under the general transformation, theoretically explaining the observation in Figure~\ref{fig:pearson}c. Addressing this remaining intra-session correlation, potentially through mechanisms that complement the general model with session-specific adaptations, presents an avenue for future research.

\ZQ{
\subsection{Limitations}
\label{sec:limitations}
This work is subject to several limitations. First, the proposed benchmark is constrained by the availability of public datasets. Despite the implementation of unified preprocessing and category mapping, the data variability is inherently limited by the existing corpus. Second, the current release focuses on conventional video streams; richer modalities such as depth are not yet integrated, and building a unified interface for these data sources remains future work. Third, while our approach addresses significant domain gaps, extreme variations, such as those stemming from distinct demographic attributes or acquisition devices, warrant further validation on more extensive and diverse datasets.
}

\ZQ{\subsection{Future Directions}
\label{sec:future_directions}
Based on the findings of this study, we propose two key directions for future research. First, addressing the issue of intra-session correlations identified in our analysis, we recommend the development of session-specific adaptation mechanisms. Integrating such mechanisms with the general evolution model could provide a more refined approach to handling complex domain shifts. Second, extending IMER from the current video-based setting to \emph{multimodal} IMER is a promising direction. While our benchmark follows a conventional video-based MER pipeline using 2D face videos, future benchmarks could integrate datasets with richer sensing modalities. Such a multimodal setting would better reflect practical deployments and may improve robustness under domain shifts. Meanwhile, multimodal IMER also introduces new challenges in (i) unifying heterogeneous data streams with different sampling rates and noise characteristics, (ii) handling missing modalities across sessions, and (iii) managing evolving label taxonomies. Future efforts should therefore focus on establishing systematic strategies for multimodal data standardization and evaluation, ensuring the sustained adaptability and robustness of MER systems.}

\section{Conclusion}
{In this paper, we established the first comprehensive benchmark for IMER, providing a systematic framework to guide future research. This benchmark formalizes the composite class-domain incremental problem unique to MER, introduces a chronologically organized data sequence, and defines robust, computationally feasible evaluation protocols.}

{Crucially, this benchmark served as a diagnostic tool. Our extensive evaluation of baseline methods revealed a fundamental stability-plasticity dilemma: existing methods either suffer from catastrophic forgetting when faced with domain shifts or lack the plasticity to adapt to new data effectively. Motivated by this core challenge, we proposed Mahalanobis Refinement (MR), a novel two-stage approach tailored for IMER. MR ensures stability by constructing a general evolution model from accumulated second-order statistics and enhances plasticity through a Mahalanobis-constrained local refinement. Our experiments demonstrate that MR not only significantly outperforms existing methods but also offers a practical and scalable solution. Together, the benchmark and the MR method provide a solid foundation and a strong starting point for advancing the field of IMER.}

\section*{Acknowledgment}

\noindent This work was funded in part by the National Natural Science Foundation of China (62376070, 62076195) and in part by the Fundamental Research Funds for the Central Universities (AUGA5710011522).

\bibliographystyle{IEEEtran}
\bibliography{references}
\end{document}